\documentclass{article} 
\usepackage{iclr2023_conference,times}

\iclrfinalcopy

\definecolor{mydarkblue}{rgb}{0,0.08,0.45}
\definecolor{myblue}{HTML}{3b75c3}
\definecolor{myred}{HTML}{E33222}
\definecolor{mygreen}{HTML}{438773}
\definecolor{mymaroon}{RGB}{142,27,19}
\definecolor{maroon}{HTML}{800000}
\definecolor{mycite}{cmyk}{0.55,1,0,0.15}
\definecolor{codeblue}{rgb}{0.25,0.5,0.5}
\definecolor{codekw}{rgb}{0.85, 0.18, 0.50}
\definecolor{codegreen}{rgb}{0,0.6,0}
\definecolor{codegray}{rgb}{0.5,0.5,0.5}
\definecolor{codepurple}{rgb}{0.58,0,0.82}
\definecolor{backcolour}{rgb}{0.95,0.95,0.92}
\definecolor{mygray}{gray}{0.925}

\usepackage{hyperref}       
\hypersetup{
    colorlinks=true,
    citecolor=mycite,
    urlcolor=mydarkblue,
    linkcolor = mydarkblue
}

\usepackage{url}
\usepackage{subcaption}
\usepackage{graphicx}
\usepackage{csquotes}
\usepackage{graphicx}
\usepackage{amsmath}
\usepackage{mathtools}
\usepackage{amssymb}
\usepackage{amsfonts}
\usepackage{bm}
\usepackage{subcaption}
\usepackage{booktabs}
\usepackage{bbm}
\usepackage{multirow}
\usepackage{enumitem}
\usepackage{wrapfig,lipsum}
\usepackage{algorithm}
\usepackage{xcolor}
\usepackage{rotating}
\usepackage{makecell}
\usepackage[capitalize,noabbrev,nameinlink]{cleveref}
\usepackage{listings}
\usepackage{colortbl}
\usepackage{pifont}
\usepackage{xspace}
\usepackage{soul}
\usepackage{etoc}

\newcolumntype{a}{>{\columncolor{mygray}}r}
\newcolumntype{b}{>{\columncolor{mygray}}c}

\newcommand{\ms}[2]{{#1\tiny{$\pm$#2}}}
\newcommand{\bms}[2]{{\textbf{#1}\tiny{$\pm$#2}}}

\makeatletter
\def\part{\par
   \addvspace{4ex}%
   \@afterindentfalse
   \secdef\@part\@spart}%
\def\@part[#1]#2{%
 \@ifnum{\c@secnumdepth >\m@ne}{%
        \refstepcounter{part}%
        \addcontentsline{toc}{part}{\thepart\hspace{1em}#1}%
 }{%
      \addcontentsline{toc}{part}{#1}%
 }%
   \nobreak
   \vskip 3ex
   \@afterheading
}%
\makeatother

\newcommand{\rstar}{\vcenter{\hbox{\includegraphics[scale=0.39]{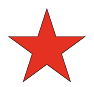}}}}
\newcommand{\bstar}{\vcenter{\hbox{\includegraphics[scale=0.39]{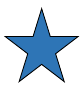}}}}
\newcommand{\gstar}{\vcenter{\hbox{\includegraphics[scale=0.39]{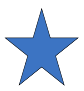}}}}

\newcommand{\mlpinit}{{\small\textsf{MLPInit}}\xspace}
\newcommand{\peermlp}{{\small\textsf{PeerMLP}}\xspace}

\newcommand{\flickr}{\texttt{Flickr}\xspace}
\newcommand{\yelp}{\texttt{Yelp}\xspace}
\newcommand{\reddit}{\texttt{Reddit}\xspace}
\newcommand{\reddittwo}{\texttt{Reddit2}\xspace}
\newcommand{\ogbnproducts}{\texttt{OGB-products}\xspace}
\newcommand{\ogbnarxiv}{\texttt{OGB-arXiv}\xspace}
\newcommand{\aproducts}{\texttt{A-products}\xspace}

\newcommand{\cora}{\texttt{Cora}\xspace}
\newcommand{\seer}{\texttt{CiteSeer}\xspace}
\newcommand{\pubmed}{\texttt{PubMed}\xspace}
\newcommand{\dblp}{\texttt{DBLP}\xspace}
\newcommand{\corafull}{\texttt{CoraFull}\xspace}
\newcommand{\photo}{\texttt{A-Photo}\xspace}
\newcommand{\computers}{\texttt{A-Computers}\xspace}
\newcommand{\cs}{\texttt{CS}\xspace}
\newcommand{\physics}{\texttt{Physics}\xspace}

\title{\textsf{MLPInit}: Embarrassingly Simple GNN Training Acceleration with MLP Initialization}

\author{Xiaotian Han\textsuperscript{1}\thanks{This work was done while the first author was an intern at Snap Inc.} \quad Tong Zhao\textsuperscript{2} \quad Yozen Liu\textsuperscript{2} \quad Xia Hu\textsuperscript{3} \quad \textbf{Neil Shah\textsuperscript{2}} \\
\textsuperscript{1}Texas A\&M University \quad \textsuperscript{2}Snap Inc. \quad \textsuperscript{3}Rice University \\
\texttt{han@tamu.edu} \quad \texttt{\{tzhao,yliu2,nshah\}@snap.com} \quad  \texttt{xia.hu@rice.edu}
}

\begin{document}

\maketitle

\begin{abstract}
Training graph neural networks (GNNs) on large graphs is complex and extremely time consuming. This is attributed to overheads caused by sparse matrix multiplication, which are sidestepped when training multi-layer perceptrons (MLPs) with only node features. MLPs, by ignoring graph context, are simple and faster for graph data, however they usually sacrifice prediction accuracy, limiting their applications for graph data. We observe that for most message passing-based GNNs, we can trivially derive an analog MLP (we call this a \peermlp) with an equivalent weight space, by setting the trainable parameters with the same shapes, making us curious about \textbf{\emph{how do GNNs using weights from a fully trained \peermlp perform?}} Surprisingly, we find that GNNs initialized with such weights significantly outperform their {\peermlp}s, motivating us to use {\peermlp} training as a precursor, initialization step to GNN training. To this end, we propose an embarrassingly simple, yet hugely effective initialization method for GNN training acceleration, called \mlpinit. Our extensive experiments on multiple large-scale graph datasets with diverse GNN architectures validate that {\mlpinit} can accelerate the training of GNNs (up to 33× speedup on \ogbnproducts) and often improve prediction performance (e.g., up to $7.97\%$ improvement for GraphSAGE across $7$ datasets for node classification, and up to $17.81\%$ improvement across $4$ datasets for link prediction on metric Hits@10). The code is available at \href{https://github.com/snap-research/MLPInit-for-GNNs}{\color{blue}{https://github.com/snap-research/MLPInit-for-GNNs}}.
\end{abstract}

\section{Introduction}\label{sec:intro}
Graph Neural Networks (GNNs) \citep{zhang2018graph,zhou2020graph,wu2020comprehensive} have attracted considerable attention from both academic and industrial researchers and have shown promising results on various practical tasks, e.g., recommendation~\citep{fan2019graph, sankar2021graph, ying2018graph, tang2022friend}, knowledge graph analysis~\citep{arora2020survey,park2019estimating,wang2021mixed}, forecasting \citep{tang2020knowing,zhao2021action,jiang2022graph} and chemistry analysis~\citep{li2018learning,you2018graphrnn,de2018molgan,liu2022graph}. However, training GNN on large-scale graphs is extremely time-consuming and costly in practice, thus spurring considerable work dedicated to scaling up the training of GNNs, even necessitating new massive graph learning libraries~\citep{zhang2020agl, ferludin2022tf} for large-scale graphs.

Recently, several approaches for more efficient GNNs training have been proposed, including novel architecture design~\citep{wu2019simplifying,you2020l2,li2021training}, data reuse and partitioning paradigms~\citep{wan2022bns,fey2021gnnautoscale,yu2022graphfm} and graph sparsification~\citep{cai2020graph,jin2021graph}. However, these kinds of methods often sacrifice prediction accuracy and increase modeling complexity, while sometimes meriting significant additional engineering efforts.

MLPs are used to accelerate GNNs~\citep{zhang2021graph,frasca2020sign,hu2021graph} by decoupling GNNs to node features learning and graph structure learning. Our work also leverages MLPs but adopts a distinct perspective. Notably, we observe that the weight space of MLPs and GNNs can be identical, which enables us to transfer weights between MLP and GNN models. Having the fact that MLPs train faster than GNNs, this observation inspired us to raise the question:
\begin{center}
\vspace{-5pt}
\textbf{\textit{Can we train GNNs more efficiently by leveraging the weights of converged MLPs?}}
\vspace{-5pt}
\end{center}
To answer this question, we first pioneer a thorough investigation to reveal the relationship between the MLPs and GNNs in terms of trainable weight space. For ease of presentation, we define the {\peermlp} of a GNN\footnote{The formal definition of \peermlp is in \cref{sec:weight_space}.} so that GNN and its {\peermlp} share the same weights~\footnote{By \textit{share the same weight}, we mean that the trainable weights of GNN and its {\peermlp} are the same in terms of size, dimension, and values.}. We find that interestingly, \emph{GNNs can be optimized by training the weights of their {\peermlp}.} Based on this observation, we adopt weights of converged {\peermlp} as the weights of corresponding GNNs and find that these GNNs perform even better than converged {\peermlp} on node classification tasks (results in \cref{tab:gnn_mlp}).

\begin{figure}[t]
    \centering
    \includegraphics[width=1.00\textwidth]{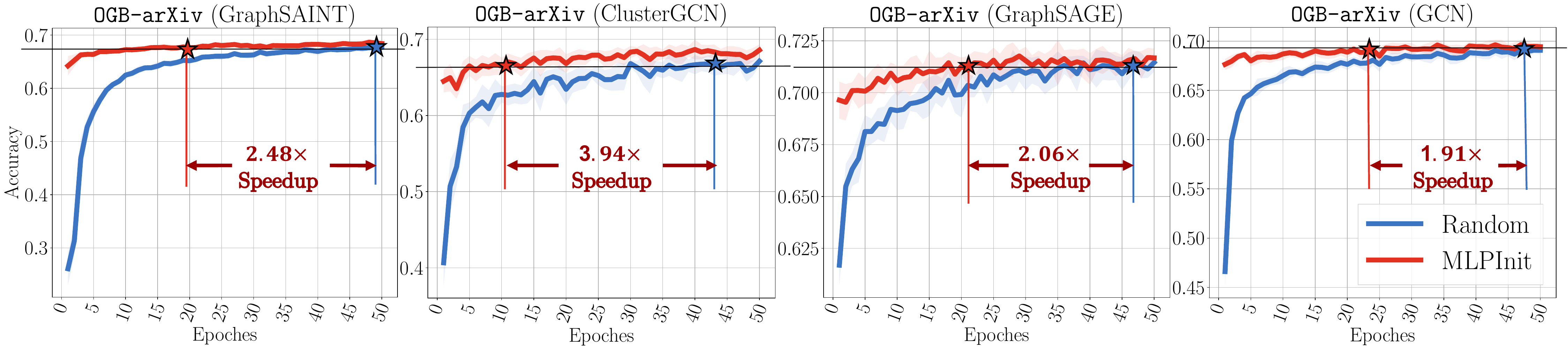}
    \vspace{-18pt}
    \caption{The training speed comparison of the GNNs with Random initialization and {\mlpinit}. $\gstar$ indicates the best performance that GNNs with random initialization can achieve. $\rstar$ indicates the comparable performance of the GNN with \mlpinit. {\color{mymaroon} Speedup} indicates the training time reduced by our proposed \mlpinit compared to random initialization. This experimental result shows that \mlpinit is able to accelerate the training of GNNs significantly.}\label{fig:intro}
    \vspace{-10pt}
\end{figure}

Motivated by this, we propose an embarrassingly simple, yet remarkably effective method to accelerate GNNs training by initializing GNN with the weights of its converged {\peermlp}. Specifically, to train a target GNN, we first train its {\peermlp} and then initialize the GNN with the optimal weights of converged {\peermlp}. We present the experimental results in \cref{fig:intro} to show the training speed comparison of GNNs with random initialization and with {\mlpinit}. In \cref{fig:intro}, {\color{mymaroon} Speedup} shows the training time reduced by our proposed {\mlpinit} compared to random initialized GNN, while achieving the same test performance. This experimental result shows that {\mlpinit} is able the accelerate the training of GNNs significantly: for example, we speed up the training of GraphSAGE, GraphSAINT, ClusterGCN, GCN by $2.48\times$, $3.94\times$, $2.06\times$, $1.91\times$ on \ogbnarxiv dataset, indicating the superiority of our method in GNNs training acceleration. Moreover, we speed up GraphSAGE training more than $14\times$ on \ogbnproducts. We highlight \textbf{our contributions} as follows:
\begin{itemize}[leftmargin=0.4cm, itemindent=.0cm, itemsep=0.0cm, topsep=0.0cm]
    \item We pioneer a thorough investigation to reveal the relationship between MLPs and GNNs in terms of the trainable weight space through the following observations: \textit{(i)} GNNs and MLPs have the same weight space. \textit{(ii)} GNNs can be optimized by training the weights of their {\peermlp}s. \textit{(iii)} GNN with weights from its converged {\peermlp} surprisingly performs better than the performance of its converged \peermlp on node classification tasks.
    \item Based on the above observations, we proposed an embarrassingly simple yet surprisingly effective initialization method to accelerate the GNNs training. Our method, called \mlpinit, initializes the weights of GNNs with the weight of their converged {\peermlp}. After initialization, we observe that GNN training takes less than half epochs to converge than those with random initialization. Thus, {\mlpinit} is able to accelerate the training of GNNs since training MLPs is cheaper and faster than training GNNs. 
    \item Comprehensive experimental results on multiple large-scale graphs with diverse GNNs validate that \mlpinit is able to accelerate the training of GNNs (up to $33\times$ speedup on \ogbnproducts) while often improving the model performance~\footnote{By \textit{performance}, we refer to the model prediction quality metric of the downstream task on the corresponding test data throughout the discussion.} (e.g., $7.97\%$ improvement for node classification on GraphSAGE and $17.81\%$ improvement for link prediction on Hits@10). 
    \item \mlpinit is extremely easy to implement and has virtually negligible computational overhead compared to the conventional GNN training schemes. In addition, it is orthogonal to other GNN acceleration methods, such as weight quantization and graph coarsening, further increasing headroom for GNN training acceleration in practice.
\end{itemize}

\section{Preliminaries}\label{sec:pre}

\textbf{Notations}. 
We denote an attributed graph $\mathcal{G} = (\mathbf{X}, \mathbf{A})$, where $\mathbf{X} = \{\mathbf{x}_1, \mathbf{x}_2,\cdots, \mathbf{x}_N\} \in \mathbb{R}^{N \times D}$ is the node feature matrix and $\mathbf{A} = \{0, 1\}^{N \times N}$ is the binary adjacency matrix. $N$ is the number of nodes, and $D$ is the dimension of node feature. For the node classification task, we denote the prediction targets by $\mathbf{Y} \in \{0, 1, \dots, C-1\}^{N}$, where $C$ is the number of classes.  We denote a GNN model as $f_{gnn}(\mathbf{X}, \mathbf{A}; w_{gnn})$, and an MLP as $f_{mlp}(\mathbf{X}; w_{mlp})$, where $w_{gnn}$ and $w_{mlp}$ denote the trainable weights in the GNN and MLP, respectively. Moreover, $w_{gnn}^{*}$ and $w_{mlp}^{*}$ denote the fixed weights of optimal (or converged) GNN and MLP, respectively.

\textbf{Graph Neural Networks}. 
Although various forms of graph neural networks (GNNs) exist, our work refers to the conventional message passing flavor \citep{gilmer2017neural}. These models work by learning a node's representation by aggregating information from the node’s neighbors recursively. One simple yet widely popular GNN instantiation is the graph convolutional network (GCN), whose multi-layer form can be written concisely: the representation vectors of all nodes at the $l$-th layer are $\mathbf{H}^{l} = \sigma( \mathbf{A} \mathbf{H}^{l-1} \mathbf{W}^{l}  )$, where $\sigma(\cdot)$ denotes activation function, $\mathbf{W}^{l}$ is the trainable weights of the $l$-th layer, and $\mathbf{H}^{l-1}$ is the node representations output by the previous layer. Denoting the output of the last layer of GNN by $\mathbf{H}$, for a node classification task, the prediction of node label is $\hat{ \mathbf{Y} } = \mathrm{softmax}( \mathbf{H} ) $. For a link prediction task, one can predict the edge probabilities with any suitable decoder, e.g., commonly used inner-product decoder as $\hat{\mathbf{A}} = \mathrm{sigmoid}( \mathbf{H}\cdot \mathbf{H}^T )$~\citep{kipf2016variational}.

\section{Motivating Analyses}\label{sec:weight_space}

In this section, we reveal that MLPs and GNNs share the same weight space, which facilitates the transferability of weights between the two architectures. Through this section, we use GCN~\citep{kipf2016semi} as a prototypical example for GNNs for notational simplicity, but we note that our discussion is generalizable to other message-passing GNN architectures.

\textbf{Motivation 1: GNNs share the same weight space with MLPs}. To show the weight space of GNNs and MLPs, we present the mathematical expression of one layer of MLP and GCN~\citep{kipf2016semi} as follows:
\begin{equation}
\label{eq:gcnmlp}
\begin{split}
\text{GNN:} \quad \mathbf{H}^{l} = \sigma( \mathbf{A} \mathbf{H}^{l-1} {\color{red} \mathbf{W}_{gnn}^{l}} ),\quad\quad\quad\quad \text{MLP:} \quad \mathbf{H}^{l} = \sigma( \mathbf{H}^{l-1} {\color{red} \mathbf{W}_{mlp}^{l}} ),
\end{split}
\end{equation}
where $\mathbf{W}_{gnn}^{l}$ and $\mathbf{W}_{mlp}^{l}$ are the trainable weights of $l$-th layer of MLP and GCN, respectively. If we set the hidden layer dimensions of GNNs and MLPs to be the same, then $\mathbf{W}_{mlp}^{l}$ and $\mathbf{W}_{gnn}^{l}$ will naturally have the same size. Thus, although the GNN and MLP are different models, their weight spaces can be identical. Moreover, for any GNN model, we can trivially derive a corresponding MLP whose weight space can be made identical. For brevity, and when the context of a GNN model is made clear, we can write such an MLP which shares the same weight space as a {\peermlp}, i.e., their trainable weights can be transferred to each other.

\textbf{Motivation 2: MLPs train faster than GNNs}. GNNs train slower than MLPs, owing to their non-trivial relational data dependency. We empirically validate that training MLPs is much faster than training GNNs in \cref{tab:runtime_mlp}. Specifically, this is because MLPs do not involve sparse matrix multiplication for neighbor aggregation. A GNN layer (here we consider a simple GCN layer, as defined in \cref{eq:gcnmlp}) can be broken down into two operations: feature transformation ($\mathbf{Z} = \mathbf{H}^{l-1} \mathbf{W}^{l}$) and neighbor aggregation ($\mathbf{H}^{l} = \mathbf{A} \mathbf{Z} $) \citep{ma2021unified}.
The neighbor aggregation and feature transformation are typically sparse and dense matrix multiplications, respectively. \cref{tab:runtime_mlp} shows the time usage for these different operations on several real-world graphs. As expected, neighbor aggregation in GNNs consumes the large majority of computation time. For example, on the \yelp dataset, the neighbor aggregation operation induces a 3199{\texttimes} time overhead.

\begin{table}[t]
\fontsize{7.5}{8}\selectfont  
\setlength{\tabcolsep}{6pt}
\centering
\caption{Comparison of the running time of forward and backward for different operations (i.e., feature transformation and neighbor aggregation) in GNNs. The time unit is milliseconds (\textit{ms}).}\label{tab:runtime_mlp}
\vspace{-8pt}
\begin{tabular}{l|rrr|rrr|rrr}
\toprule
\multicolumn{1}{c}{Operation}       & \multicolumn{3}{c}{\ogbnarxiv}            & \multicolumn{3}{c}{\flickr}                & \multicolumn{3}{c}{\yelp}     \\    \midrule
\multicolumn{1}{c}{\#Nodes}          & \multicolumn{3}{c}{169343}                & \multicolumn{3}{c}{89250}                 & \multicolumn{3}{c}{716847}     \\   
\multicolumn{1}{c}{\#Edges}          & \multicolumn{3}{c}{1166243}               & \multicolumn{3}{c}{899756}                & \multicolumn{3}{c}{13954819}     \\
\midrule
                                    &Forward     & Backward   &Total            &Forward     & Backward   &Total            &Forward     & Backward     &Total    \\    \midrule
                      $Z = WX$      &0.32     & 1.09    &1.42          &0.28     & 0.97    &1.26          &1.58     & 4.41      &5.99      \\
                      $H = AZ$      &1.09     & 1028.08 &1029.17       &1.01     & 836.95  &837.97        &9.74     & 19157.17  &19166.90     \\   
                                    &            &            &$724\times$      &            &            &$665\times$      &            &              &$3199\times$     \\   
                      \bottomrule
\end{tabular}
\end{table}

Given that the weights of GNNs and their {\peermlp} can be transferred to each other, but the {\peermlp} can be trained much faster, we raise the following questions:

\begin{enumerate}[itemsep=0.0cm, topsep=0.0cm]
    \item \textbf{\textit{What will happen if we directly adopt the weights of a converged {\peermlp} to GNN? }}
    \item \textbf{\textit{To what extent can \peermlp speed up GNN training and improve GNN performance?}}
\end{enumerate}

In this paper, we try to answer these questions with a comprehensive empirical analysis.

\section{What will happen if we directly adopt the weights of a converged \peermlp to GNN?}\label{sec:what_will_happen}

\begin{figure}[t]
    \centering
    \vspace{-10pt}
    \includegraphics[width=0.99\textwidth]{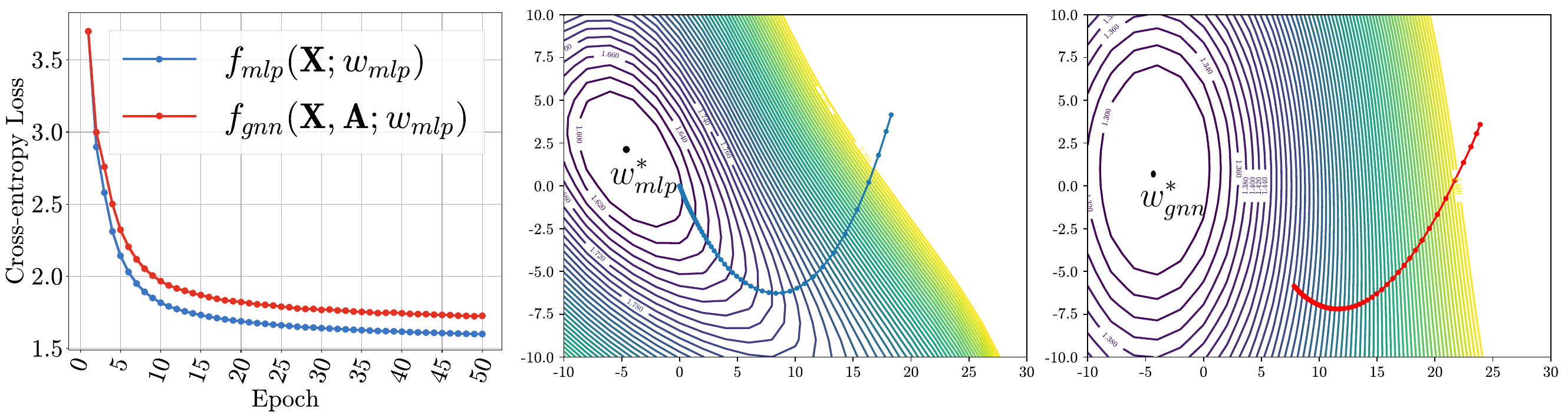}
    \vspace{-8pt}
    \caption{The relation of GNN and MLP during the training of PeerMLP. \textbf{Left}: Cross-Entropy loss of $f_{gnn}(\mathbf{X}, \mathbf{A}; w_{mlp})$ ({\color{myred} GNN}) and $f_{mlp}(\mathbf{X}; w_{mlp})$ ({\color{myblue} PeerMLP}) on training set over training epochs of PeerMLP. In this experiment, GNN and PeerMLP share the same weight $w_{mlp}$, which are trained by the PeerMLP. \textbf{Middle}: training trajectory of PeerMLP on its own loss landscape. \textbf{Right}: training trajectory of GNN with weights from PeerMLP on GNN's loss landscape. The figures show that training loss of GNN with weights trained from MLP will decrease. The details are presented in \cref{sec:app:imple:fig:gnn_mlp_loss}. We also present loss curves on validation/test sets and accuracy curves in \cref{sec:app:addition:loss_acc}.}\label{fig:gnn_mlp_loss}
    \vspace{-10pt}
\end{figure}

To answer this question, we conducted comprehensive preliminary experiments to investigate weight transferability between MLPs and GNNs. We made the following interesting and inspiring findings:

\textbf{Observation 1: The training loss of GNN will decrease by optimizing the weights of its {\peermlp}}. 
We conducted a verification experiment to investigate the loss changes of the GNNs with the weights trained from its {\peermlp} and present the results in \cref{fig:gnn_mlp_loss}. In this experiment, we have two models, a {\color{myred}GNN} and its corresponding {\color{myblue}{\peermlp}}, who share the same weights $w_{mlp}$. That is, the {\peermlp} is $f_{mlp}(\mathbf{X}; w_{mlp})$ and the GNN is $f_{gnn}(\mathbf{X},\mathbf{A}; w_{mlp})$. We optimize the weights $w_{mlp}$ by training the {\peermlp}, and the loss curve of $f_{mlp}(\mathbf{X}; w_{mlp})$ is the {\color{myblue} blue} line in the left figure in \cref{fig:gnn_mlp_loss}. We also compute the loss of GNN $f_{gnn}(\mathbf{X},\mathbf{A}; w_{mlp})$ with the weights from {\peermlp}. The loss curve of $f_{gnn}(\mathbf{X},\mathbf{A}; w_{mlp})$ is shown in the {\color{myred} red} line. \cref{fig:gnn_mlp_loss} shows the surprising phenomenon that the training loss of GNN with weights trained from {\peermlp} decreases consistently. Impressively, these weights ($w_{mlp}$) were derived without employing neighbor aggregation in training. 

\begin{wraptable}[13]{R}{0.48\textwidth}
\vspace{-10pt}
\centering
\fontsize{6.9}{9.5}\selectfont
\setlength{\tabcolsep}{2pt}
\caption{The performance of GNNs and its {\peermlp} with the weights of a converged {\peermlp} on test data.}\label{tab:gnn_mlp}
\vspace{-8pt}
\begin{tabular}{l|cccc|c} 
\toprule
                                                                                &Methods     & {\textsf{PeerMLP}}          & GNN                      & Improv.                           &MLPInit  \\ \midrule
\multirow{4}{*}{\rotatebox{90}{{\fontsize{6}{4}\selectfont \ogbnarxiv}}}        &GraphSAGE   & \ms{56.04}{0.27}    &\ms{62.87}{0.95}    	   &{\color{blue}$\uparrow 12.18\%$}       &\ms{72.25}{0.30} \\
                                                                                &GraphSAINT  & \ms{53.88}{0.41}    &\ms{63.26}{0.71}    	   &{\color{blue}$\uparrow 17.41\%$}       &\ms{68.80}{0.20} \\
                                                                                &ClusterGCN  & \ms{54.47}{0.41}    &\ms{60.81}{1.30}    	   &{\color{blue}$\uparrow 11.63\%$}       &\ms{69.53}{0.50} \\ 
                                                                                &GCN         & \ms{56.31}{0.21}    &\ms{56.28}{0.89}    	   &{\color{myred}$\downarrow 0.04\%$}      &\ms{70.35}{0.34} \\ \midrule
\multirow{4}{*}{\rotatebox{90}{{\fontsize{6}{4}\selectfont  \ogbnproducts}}}    &GraphSAGE   & \ms{63.43}{0.14}    &\ms{74.32}{1.04}    	   &{\color{blue}$\uparrow 17.16\%$}       &\ms{80.04}{0.62} \\
                                                                                &GraphSAINT  & \ms{57.29}{0.32}    &\ms{69.00}{1.54}    	   &{\color{blue}$\uparrow 20.44\%$}       &\ms{74.02}{0.19} \\ 
                                                                                &ClusterGCN  & \ms{59.53}{0.46}    &\ms{71.74}{0.70}    	   &{\color{blue}$\uparrow 20.51\%$}       &\ms{78.48}{0.64} \\ 
                                                                                &GCN         & \ms{62.63}{0.15}    &\ms{71.11}{0.10}    	   &{\color{blue}$\uparrow 13.55\%$}       &\ms{76.85}{0.34}\\ 
\bottomrule
\end{tabular}
\end{wraptable}
\textbf{Observation 2: Converged weights from {\peermlp} provide a good GNN initialization.} As {\peermlp} and GNN have the same weight spaces, a natural follow-up question is whether GNN can directly adopt the weights of the converged {\peermlp} and perform well. We next aim to understand this question empirically. Specifically, we first trained a {\peermlp} for a target GNN and obtained the optimal weights $w_{mlp}^{*}$. Next, we run inference on test data using a GNN with $w_{mlp}^{*}$ of {\peermlp}, i.e., applying $f_{gnn}(\mathbf{X},\mathbf{A}; w_{mlp}^{*})$.
\cref{tab:gnn_mlp} shows the results of $f_{mlp}(\mathbf{X}; w_{mlp}^{*})$ and $f_{gnn}(\mathbf{X},\mathbf{A}; w_{mlp}^{*})$. We can observe that the GNNs with the optimal weights of {\peermlp} consistently outperform {\peermlp}, indicating that the weights from converged \peermlp can serve as good enough initialization of the weights of GNNs.

\subsection{The Proposed Method: {\mlpinit}}\label{sec:what_will_happen:the_proposed}

The above findings show that MLPs can help the training of GNNs. In this section, we formally present our method {\mlpinit}, which is an embarrassingly simple, yet extremely effective approach to accelerating GNN training.

The basic idea of {\mlpinit} is straightforward: we adopt the weights of a converged {\peermlp} to initialize the GNN, subsequently, fine-tune the GNN. Specifically, for a target GNN ($f_{gnn}(\mathbf{X},\mathbf{A}; w_{gnn})$), we first construct a {\peermlp} ($f_{mlp}(\mathbf{X},\mathbf{A}; w_{mlp})$), with matching target weights. Next, we optimize the weight of the {\peermlp} model by training the {\peermlp} solely with the node features $\mathbf{X}$ for $m$ epochs. Upon training the {\peermlp} to convergence and obtaining the optimal weights ($w_{mlp}^{*}$), we initialize the GNN with $w_{mlp}^{*}$ and then fine-tune the GNN with $n$ epochs. We present PyTorch-style pseudo-code of \mlpinit in node classification setting in \cref{alg:mlpinit}.

 \begin{wrapfigure}[17]{R}{0.50\textwidth}
    \begin{minipage}{0.50\textwidth}
    \vspace{-0.3in}
    \begin{algorithm}[H]
    \caption{\footnotesize PyTorch-style Pseudocode of {\mlpinit}}\label{alg:mlpinit}
  \begin{lstlisting}[language=python]
  # f_gnn: graph neural network model
  # f_mlp: PeerMLP of f_gnn

  # Train PeerMLP for N epochs
  for X, Y in dataloader_mlp:
      P = f_mlp(X)
      loss = nn.CrossEntropyLoss(P, Y)
      loss.backward()
      optimizer_mlp.step()
  
  # Initialize GNN with MLPInit
  torch.save(f_mlp.state_dict(), "w_mlp.pt")
  f_gnn.load_state_dict("w_mlp.pt")
  
  # Train GNN for n epochs
  for X, A, Y in dataloader_gnn:
      P = f_gnn(X, A)
      loss = nn.CrossEntropyLoss(P, Y)
      loss.backward()
      optimizer_gnn.step()
  \end{lstlisting}
    \end{algorithm}
    \end{minipage}
  \end{wrapfigure}
  
\textbf{Training Acceleration}. Since training of the {\peermlp} is comparatively cheap, and the weights of the converged {\peermlp} can provide a good initialization for the corresponding GNN, the end result is that we can significantly reduce the training time of the GNN. Assuming that the training of GNN from a random initialization needs $N$ epochs to converge, and $N >> n$, the total training time can be largely reduced given that MLP training time is negligible compared to GNN training time. The experimental results in \cref{tab:speed} show that $N$ is generally much larger than $n$.

\textbf{Ease of Implementation}. \mlpinit is extremely easy to implement as shown in \cref{alg:mlpinit}. First, we construct an MLP ({\peermlp}), which has the same weights with the target GNN. Next, we use the node features $\mathbf{X}$ and node labels $\mathbf{Y}$ to train the {\peermlp} to converge. Then, we adopt the weights of converged {\peermlp} to the GNN, and fine-tune the GNN while additionally leveraging the adjacency $\mathbf{A}$. In addition, our method can also directly serve as the final, or deployed GNN model, in resource-constrained settings: assuming $n = 0$, we can simply train the {\peermlp} and adopt $w^*_{mlp}$ directly. This reduces training cost further, while enabling us to serve a likely higher performance model in deployment or test settings, as \cref{tab:gnn_mlp} shows.

\subsection{Discussion}\label{sec:meth:dissuss}
In this section, we discuss the relation between \mlpinit and existing methods. Since we position \mlpinit as an acceleration method involved MLP, we first compare it with MLP-based GNN acceleration methods, and we also compare it with GNN Pre-training methods.

\textbf{Comparison to MLP-based GNN Acceleration Methods}.
Recently, several works aim to simplify GNN to MLP-based constructs during training or inference \citep{zhang2022graph,wu2019simplifying,frasca2020sign,sun2021scalable,huang2020combining,hu2021graph}. Our method is proposed to accelerate the message passing based GNN for large-scale graphs. Thus, MLP-based GNN acceleration is a completely different line of work compared to ours since it removes the message passing in the GNNs and uses MLP to model graph structure instead. Thus, MLP-based GNN acceleration methods are out of the scope of the discussion in this work.

\textbf{Comparison to GNN Pre-training Methods}.
Our proposed \mlpinit are orthogonal to the GNN pre-training methods\citep{you2020graph,zhu2020deep,velivckovic2018deep,you2021graph,qiu2020gcc,zhu2021graph,hu2019strategies}. GNN pre-training typically leverages graph augmentation to pretrain weights of GNNs or obtain the node representation for downstream tasks. Compared with the pre-training methods, \mlpinit has two main differences (or advantages) that significantly contribute to the speed up: (i) the training of \peermlp does not involve using the graph structure data, while pre-training methods rely on it. (ii) Pre-training methods usually involve graph data augmentation~\citep{qiu2020gcc,zhao2022graph}, which requires additional training time.

\section{Experiments}
In the next subsections, we conduct and discuss experiments to understand \mlpinit from the following aspects: \textbf{(i)} training speedup, \textbf{(ii)} performance improvements, \textbf{(iii)} hyperparameter sensitivity, \textbf{(iv)} robustness and loss landscape. For node classification, we consider \flickr, \yelp, \reddit, \reddittwo, \aproducts, and two OGB datasets \citep{hu2020open}, \ogbnarxiv and \ogbnproducts as benchmark datasets. We adopt {GCN} (w/ mini-batch) \citep{kipf2016semi}, {GraphSAGE} \citep{hamilton2017inductive}, GraphSAINT\citep{zeng2019graphsaint} and ClusterGCN \citep{chiang2019cluster} as GNN backbones. The details of datasets and baselines are in \cref{sec:app:setting:data,sec:app:setting:base}, respectively. For the link prediction task, we consider \cora, \seer, \pubmed, \corafull, \cs, \physics, \photo, and \computers as our datasets. Our link prediction setup is using as GCN as an encoder which transforms a graph to node representation $\mathbf{H}$ and an inner-product decoder $\hat{\mathbf{A}} = \mathrm{sigmoid}( \mathbf{H}\cdot \mathbf{H}^T )$ to predict the probability of the link existence, which is discussed in \cref{sec:pre}.

\subsection{How much can {\mlpinit} accelerate GNN training?}\label{sec:exp:accelerate}
\begin{table}[t]
    \fontsize{9}{9}\selectfont
    \setlength{\tabcolsep}{5pt}
    \centering
    \caption{Speed improvement when {\mlpinit} achieves comparable performance with random initialized GNN. The number reported is the training epochs needed. ({\color{blue}---}) means our method can not reach comparable performance. The epoch used by Random/\mlpinit is denoted as $\bstar$/$\rstar$ in \cref{fig:intro}. The detailed speedup computation method are presented in \cref{sec:app:imple:fig:ogbn_products_perf}.
    }\label{tab:speed}
    \vspace{-8pt}
    \begin{tabular}{clrrrrrrra}
    \toprule
                            &Methods    & {\scriptsize\flickr} & {\scriptsize\yelp} & {\scriptsize\reddit} & {\scriptsize\reddittwo}  & {\scriptsize\aproducts} & {\scriptsize\ogbnarxiv}       & {\scriptsize\ogbnproducts}       &Avg.             \\\midrule
\multirow{3}{*}{\rotatebox{90}{SAGE}}   &Random($\bstar$)       &45.6		                    &44.7		                    &36.0		                    &48.0		                    &48.9		                    &46.7		                        &43.0		                    &44.7\\
                                        &\mlpinit($\rstar$)     &39.9		                    &20.3		                    &7.3		                    &7.7		                    &40.8		                    &22.7		                        &2.9		                    &20.22\\
                                        &Improv.                &{\color{blue}1.14$\times$}      &{\color{blue}2.20$\times$}      &{\color{blue}4.93$\times$}      &{\color{blue}6.23$\times$}      &{\color{blue}1.20$\times$}      &{\color{blue}2.06$\times$}          &{\color{blue}14.83$\times$}     &{\color{blue}2.21$\times$} \\\midrule
\multirow{3}{*}{\rotatebox{90}{SAINT}}  &Random                 &31.0		                    &35.8		                    &40.6		                    &28.3		                    &50.0		                    &48.3		                        &44.9		                    &40.51\\
                                        &\mlpinit               &14.1		                    &0.0		                    &21.8		                    &6.1		                    &9.1		                    &19.5		                        &16.9		                    &{14.58}\\
                                        &Improv.                &{\color{blue}2.20$\times$}      &{\color{blue}---}               &{\color{blue}1.86$\times$}      &{\color{blue}4.64$\times$}      &{\color{blue}5.49$\times$}      &{\color{blue}2.48$\times$}          &{\color{blue}2.66$\times$}      &{\color{blue}2.77$\times$} \\\midrule
\multirow{3}{*}{\rotatebox{90}{C-GCN}}  &Random                 &15.7		                    &40.3		                    &46.2		                    &47.0		                    &37.4		                    &42.9		                        &42.8		                    &38.9\\
                                        &\mlpinit               &7.3		                    &18.0		                    &12.8		                    &17.0		                    &1.0		                    &10.9		                        &15.0		                    &11.7\\
                                        &Improv.                &{\color{blue}2.15$\times$}      &{\color{blue}2.24$\times$}      &{\color{blue}3.61$\times$}      &{\color{blue}2.76$\times$}      &{\color{blue}37.40$\times$}     &{\color{blue}3.94$\times$}          &{\color{blue}2.85$\times$}      &{\color{blue}3.32$\times$} \\\midrule
\multirow{3}{*}{\rotatebox{90}{GCN}}    &Random                 &46.4		                    &44.5		                    &42.4		                    &2.4		                    &47.7		                    &46.7		                        &43.8		                    &{45.35}\\
                                        &\mlpinit               &30.5		                    &23.3		                    &0.0		                    &0.0		                    &0.0		                    &24.5		                        &1.3		                    &{19.9}\\
                                        &Improv.                &{\color{blue}1.52$\times$}      &{\color{blue}1.91$\times$}      &{\color{blue}---}               &{\color{blue}---}               &{\color{blue}---}               &{\color{blue}1.91$\times$}          &{\color{blue}33.69$\times$}     &{\color{blue}2.27$\times$}    \\
        
    \bottomrule
    \end{tabular}
\end{table}

\begin{figure}[t]
    \centering
    \vspace{-8pt}
    \includegraphics[width=0.99\textwidth]{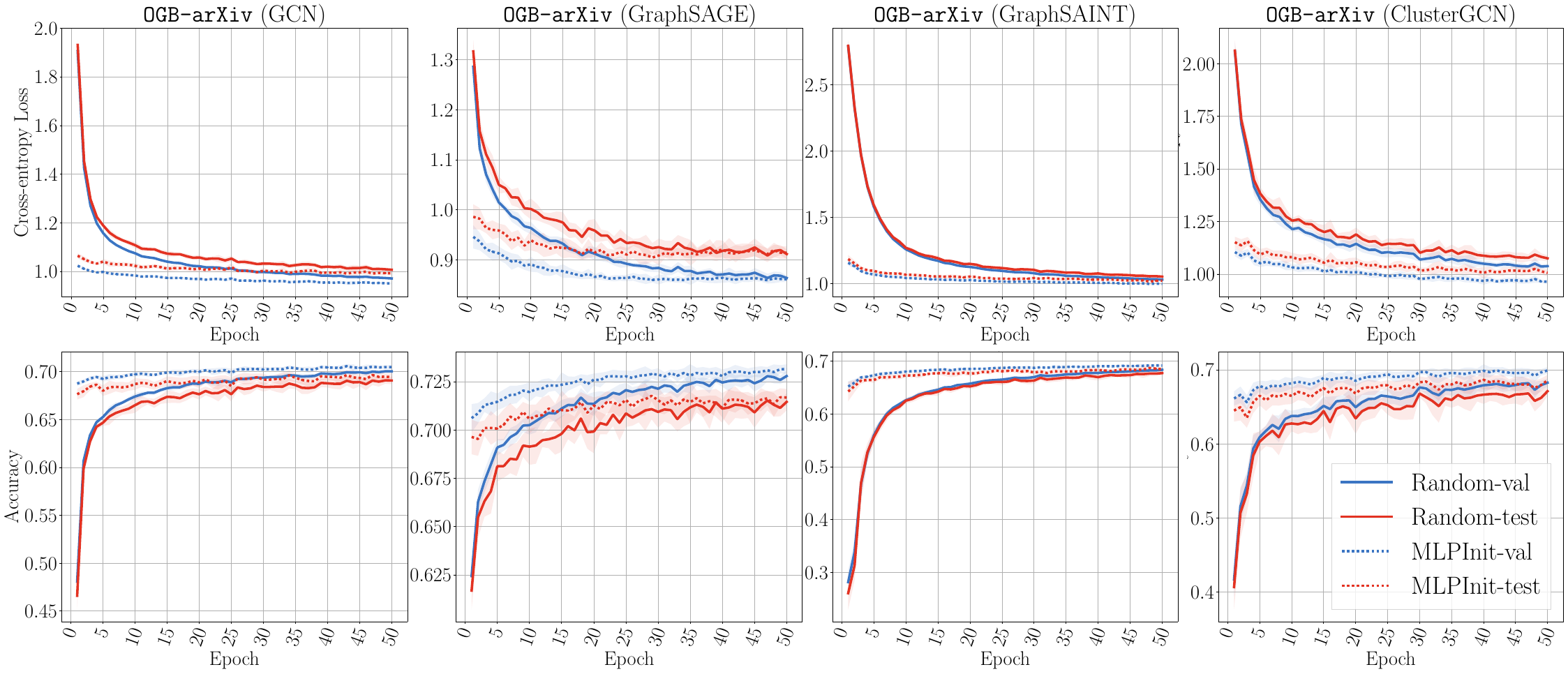}
    \vspace{-8pt}
    \caption{The training curves of different GNNs on \ogbnarxiv. GNN with \mlpinit generally obtains lower loss and higher accuracy than those with the random initialization and converges faster. The training curves are depicted based on ten runs. More experiment results are in \cref{sec:app:add}.}\label{fig:ogbn_arxiv_perf}
    \vspace{-8pt}
\end{figure}

In this section, we compared the training speed of GNNs with random initialization and {\mlpinit}. We computed training epochs needed by GNNs with random initialization to achieve the best test performance. We also compute the running epochs needed by  GNNs with \mlpinit to achieve comparable test performance. We present the results in \cref{tab:speed}. We also plotted the loss and accuracy curves of different GNNs on \ogbnarxiv in \cref{fig:ogbn_arxiv_perf}. We made the following major observations:

\textbf{Observation 3: {\mlpinit} can significantly reduce the training time of GNNs. }  In this experiment, we summarize the epochs needed by GNN with random initialization to obtain the best performance, and then we calculate the epochs needed by GNN with \mlpinit to reach a comparable performance on par with the randomly initialized GNN. We present the time speedup of {\mlpinit} in \cref{tab:speed}. \cref{tab:speed} shows \mlpinit speed up the training of GNNs by $2-5$ times generally and in some cases even more than $30$ times. The consistent reduction of training epochs on different datasets demonstrates that \mlpinit can generally speed up GNN training quite significantly.

\subsection{How well does {\mlpinit} perform on node classification and link prediction tasks?}\label{sec:exp:perform_nc}
In this section, we conducted experiments to show the superiority of the proposed method in terms of final, converged GNN model performance on node classification and link prediction tasks. The reported test performances of both random initialization and \mlpinit are selected based on the validation data. We present the performance improvement of {\mlpinit} compared to random initialization in \cref{tab:perf,tab:lp} for node classification and link prediction, respectively. 
\begin{table}[t]
    \fontsize{9}{10}\selectfont
    \setlength{\tabcolsep}{2.7pt}
    \centering
    \caption{Performance improvement when GNN with random initialization and with \mlpinit achieve best test performance, respectively. Mean and standard deviation are calculated based on ten runs. The best test performance for the two methods is independently selected based on validation data.}\label{tab:perf}
    \vspace{-8pt}
    \begin{tabular}{clrrrrrrra}
    \toprule
                                        &Methods            & {\scriptsize\flickr}              & {\scriptsize\yelp}              & {\scriptsize\reddit}            & {\scriptsize\reddittwo}           & {\scriptsize\aproducts}           & {\scriptsize\ogbnarxiv}        & {\scriptsize\ogbnproducts} &Avg. \\\midrule
\multirow{3}{*}{\rotatebox{90}{SAGE}}   &Random             &\ms{53.72}{0.16}			        &\ms{63.03}{0.20}			      &\ms{96.50}{0.03}			        &\ms{51.76}{2.53}			        &\ms{77.58}{0.05}			        &\ms{72.00}{0.16}			     &\ms{80.05}{0.35}			            &70.66\\ 
                                        &\mlpinit           &\ms{53.82}{0.13}			        &\ms{63.93}{0.23}			      &\ms{96.66}{0.04}			        &\ms{89.60}{1.60}			        &\ms{77.74}{0.06}			        &\ms{72.25}{0.30}			     &\ms{80.04}{0.62}			            &76.29\\
                                        &Improv.            &{\color{blue}$\uparrow 0.19\%$}     &{\color{blue}$\uparrow 1.43\%$}   &{\color{blue}$\uparrow 0.16\%$}   &{\color{blue}$\uparrow 73.09\%$}    &{\color{blue}$\uparrow 0.21\%$}     &{\color{blue}$\uparrow 0.36\%$}  &{\color{myred}$\downarrow 0.01\%$}	    &{\color{blue}$\uparrow 7.97\%$}\\ \midrule
\multirow{3}{*}{\rotatebox{90}{SAINT}}  &Random             &\ms{51.37}{0.21}			        &\ms{29.42}{1.32}			      &\ms{95.58}{0.07}			        &\ms{36.45}{4.09}			        &\ms{59.31}{0.12}			        &\ms{67.95}{0.24}			     &\ms{73.80}{0.58}			            &59.12\\
                                        &\mlpinit          &\ms{51.35}{0.10}			        &\ms{43.10}{1.13}			      &\ms{95.64}{0.06}			        &\ms{41.71}{1.25}			        &\ms{68.24}{0.17}			        &\ms{68.80}{0.20}			     &\ms{74.02}{0.19}			            &63.26\\
                                        &Improv.            &{\color{myred}$\downarrow 0.05\%$}  &{\color{blue}$\uparrow 46.47\%$}  &{\color{blue}$\uparrow 0.06\%$}   &{\color{blue}$\uparrow 14.45\%$}    &{\color{blue}$\uparrow 15.06\%$}    &{\color{blue}$\uparrow 1.25\%$}  &{\color{blue}$\uparrow 0.30\%$}	    &{\color{blue}$\uparrow 7.00\%$}\\\midrule
\multirow{3}{*}{\rotatebox{90}{C-GCN}}  &Random             &\ms{49.95}{0.15}			        &\ms{56.39}{0.64}			      &\ms{95.70}{0.06}			        &\ms{53.79}{2.48}			        &\ms{52.74}{0.28}			        &\ms{68.00}{0.59}			     &\ms{78.71}{0.59}			            &65.04\\
                                        &\mlpinit          &\ms{49.96}{0.20}			        &\ms{58.05}{0.56}			      &\ms{96.02}{0.04}			        &\ms{77.77}{1.93}			        &\ms{55.61}{0.17}			        &\ms{69.53}{0.50}			     &\ms{78.48}{0.64}			            &69.34\\
                                        &Improv.            &{\color{blue}$\uparrow 0.02\%$}     &{\color{blue}$\uparrow 2.94\%$}   &{\color{blue}$\uparrow 0.33\%$}   &{\color{blue}$\uparrow 44.60\%$}    &{\color{blue}$\uparrow 5.45\%$}     &{\color{blue}$\uparrow 2.26\%$}  &{\color{myred}$\downarrow 0.30\%$}	    &{\color{blue}$\uparrow 6.61\%$}\\\midrule
\multirow{3}{*}{\rotatebox{90}{GCN}}    &Random             &\ms{50.90}{0.12}			        &\ms{40.08}{0.15}			      &\ms{92.78}{0.11}			        &\ms{27.87}{3.45}			        &\ms{36.35}{0.15}			        &\ms{70.25}{0.22}			     &\ms{77.08}{0.26}			            &56.47\\
                                        &\mlpinit          &\ms{51.16}{0.20}			        &\ms{40.83}{0.27}			      &\ms{91.40}{0.20}			        &\ms{80.37}{2.61}			        &\ms{39.70}{0.11}			        &\ms{70.35}{0.34}			     &\ms{76.85}{0.34}			            &64.38\\
                                        &Improv.            &{\color{blue}$\uparrow 0.51\%$}     &{\color{blue}$\uparrow 1.87\%$}   &{\color{myred}$\downarrow 1.49\%$}  &{\color{blue}$\uparrow 188.42\%$}   &{\color{blue}$\uparrow 9.22\%$}     &{\color{blue}$\uparrow 0.14\%$}  &{\color{myred}$\downarrow 0.29\%$}    &{\color{blue}$\uparrow 14.00\%$}\\
    \bottomrule
    \end{tabular}
\end{table}

\begin{table}[t]
\centering
\fontsize{9}{9}\selectfont
\setlength{\tabcolsep}{7.9pt}
\vspace{-8pt}
\caption{The performance of link prediction task. The results are based on ten runs. The experiments on other datasets are presented in \cref{tab:lp_app}. More experiments are presented in \cref{sec:app:add:lp}.}\label{tab:lp}
\vspace{-8pt}
\begin{tabular}{l|l|rrrrrrr} 
\toprule
                            &Methods                                &AUC                &AP                 & Hits@10           & Hits@20           & Hits@50            & Hits@100 \\ \midrule

\multirow{4}{*}{\rotatebox{90}{\scriptsize\pubmed}}     & $\textrm{MLP}_{\textrm{random}}$      &\ms{94.76} {0.30}	&\ms{94.28} {0.36}	&\ms{14.68} {2.60}	&\ms{24.01} {3.04}	&\ms{40.02} {2.75}	&\ms{54.85} {2.03} \\
                            & $\textrm{GNN}_{\textrm{random}}$      &\ms{96.66} {0.29}	&\ms{96.78} {0.31}	&\ms{28.38} {6.11}	&\ms{42.55} {4.83}	&\ms{60.62} {4.29}	&\ms{75.14} {3.00} \\
                            & $\textrm{GNN}_{\textrm{mlpinit}}$     &\ms{97.31} {0.19}	&\ms{97.53} {0.21}	&\ms{37.58} {7.52}	&\ms{51.83} {7.62}	&\ms{70.57} {3.12}	&\ms{81.42} {1.52} \\
                            & Improvement &{\color{blue}$\uparrow 0.68\%$}	&{\color{blue}$\uparrow 0.77\%$}	&{\color{blue}$\uparrow 32.43\%$}	 &{\color{blue}$\uparrow 21.80\%$}	&{\color{blue}$\uparrow 16.42\%$}	&{\color{blue}$\uparrow 8.36\%$}\\ \midrule

\multirow{4}{*}{\rotatebox{90}{\scriptsize\dblp}}       & $\textrm{MLP}_{\textrm{random}}$      &\ms{95.20} {0.18}	&\ms{95.53} {0.25}	&\ms{28.70} {3.73}	&\ms{39.22} {4.13}	&\ms{53.36} {3.81}	&\ms{64.83} {1.95} \\
                            & $\textrm{GNN}_{\textrm{random}}$      &\ms{96.29} {0.20}	&\ms{96.64} {0.23}	&\ms{36.55} {4.08}	&\ms{43.13} {2.85}	&\ms{59.98} {2.43}	&\ms{71.57} {1.00} \\
                            & $\textrm{GNN}_{\textrm{mlpinit}}$     &\ms{96.67} {0.13}	&\ms{97.09} {0.14}	&\ms{40.84} {7.34}	&\ms{53.72} {4.25}	&\ms{67.99} {2.85}	&\ms{77.76} {1.20} \\
                            & Improvement      &{\color{blue}$\uparrow 0.39\%$}	&{\color{blue}$\uparrow 0.47\%$}	&{\color{blue}$\uparrow 11.73\%$}	&{\color{blue}$\uparrow 24.57\%$}	&{\color{blue}$\uparrow 13.34\%$}	&{\color{blue}$\uparrow 8.65\%$}     \\ \midrule
                            
\multirow{4}{*}{\rotatebox{90}{\scriptsize\photo}}      & $\textrm{MLP}_{\textrm{random}}$      &\ms{86.18} {1.41}	&\ms{85.37} {1.24}	&\ms{4.36} {1.14}	&\ms{6.96} {1.28}	&\ms{12.20} {1.24}	&\ms{17.91} {1.26} \\
                            & $\textrm{GNN}_{\textrm{random}}$      &\ms{92.07} {2.14}	&\ms{91.52} {2.08}	&\ms{9.63} {1.58}	&\ms{12.82} {1.72}	&\ms{20.90} {1.90}	&\ms{29.08} {2.53} \\
                            & $\textrm{GNN}_{\textrm{mlpinit}}$     &\ms{93.99} {0.58}	&\ms{93.32} {0.60}	&\ms{9.17} {2.12}	&\ms{13.12} {2.11}	&\ms{22.93} {2.56}	&\ms{32.37} {1.89} \\
                            & Improvement     &{\color{blue}$\uparrow 2.08\%$}	&{\color{blue}$\uparrow 1.97\%$}	&{\color{myred}$\downarrow 4.75\%$}	&{\color{blue}$\uparrow 2.28\%$}	&{\color{blue}$\uparrow 9.73\%$}	&{\color{blue}$\uparrow 11.32\%$}              \\ \midrule
                            
\multirow{4}{*}{\rotatebox{90}{\scriptsize\physics}}    & $\textrm{MLP}_{\textrm{random}}$      &\ms{96.26} {0.11}	&\ms{95.63} {0.15}	&\ms{5.38} {1.32}	&\ms{8.76} {1.37}	&\ms{15.86} {0.81}	&\ms{24.70} {1.11} \\
                            & $\textrm{GNN}_{\textrm{random}}$      &\ms{95.84} {0.13}	&\ms{95.38} {0.15}	&\ms{6.62} {1.00}	&\ms{10.39} {1.04}	&\ms{18.55} {1.60}	&\ms{26.88} {1.95} \\
                            & $\textrm{GNN}_{\textrm{mlpinit}}$     &\ms{96.89} {0.07}	&\ms{96.55} {0.11}	&\ms{8.05} {1.44}	&\ms{13.06} {1.94}	&\ms{22.38} {1.94}	&\ms{32.31} {1.43} \\
                            & Improvement     &{\color{blue}$\uparrow 1.10\%$}	&{\color{blue}$\uparrow 1.22\%$}	&{\color{blue}$\uparrow 21.63\%$}	&{\color{blue}$\uparrow 25.76\%$}	&{\color{blue}$\uparrow 20.63\%$}	&{\color{blue}$\uparrow 20.20\%$}    \\ \midrule
\rowcolor{mygray}
\multicolumn{2}{c}{Avg.}    &{\color{blue}$\uparrow 1.05\%$}	&{\color{blue}$\uparrow 1.10\%$}	&{\color{blue}$\uparrow 17.81\%$}	&{\color{blue}$\uparrow 20.97\%$}	&{\color{blue}$\uparrow 14.88\%$}	&{\color{blue}$\uparrow 10.46\%$} \\

\bottomrule
\end{tabular}
\vspace{-8pt}
\end{table}

\textbf{Observation 4: {\mlpinit} improves the prediction performance for both node classification and link prediction task in most cases.}  \cref{tab:perf} shows our proposed method gains $7.97\%$, $7.00\%$, $6.61\%$ and $14.00\%$ improvements for GraphSAGE, GraphSAINT, ClusterGCN, and GCN on average cross all the datasets for the node classification task. The results in \cref{tab:lp} and \cref{tab:lp_app} show our proposed method gains $1.05\%$, $1.10\%$, $17.81\%$, $20.97\%$, $14.88\%$,$10.46\%$ on average cross various metrics for the link prediction task.

\subsection{Is {\mlpinit} robust under different hyperparameters?}

In practice, one of the most time-consuming parts of training large-scale GNNs is hyperparameter tuning \citep{you2020design}. Here, we perform experiments to investigate the sensitivity of {\mlpinit} to various hyperparameters, including the architecture hyperparameters and training hyperparameters.

\begin{wrapfigure}[18]{r}{0.45\textwidth}
    \centering
    \vspace{-12pt}
    \includegraphics[width=0.45\textwidth]{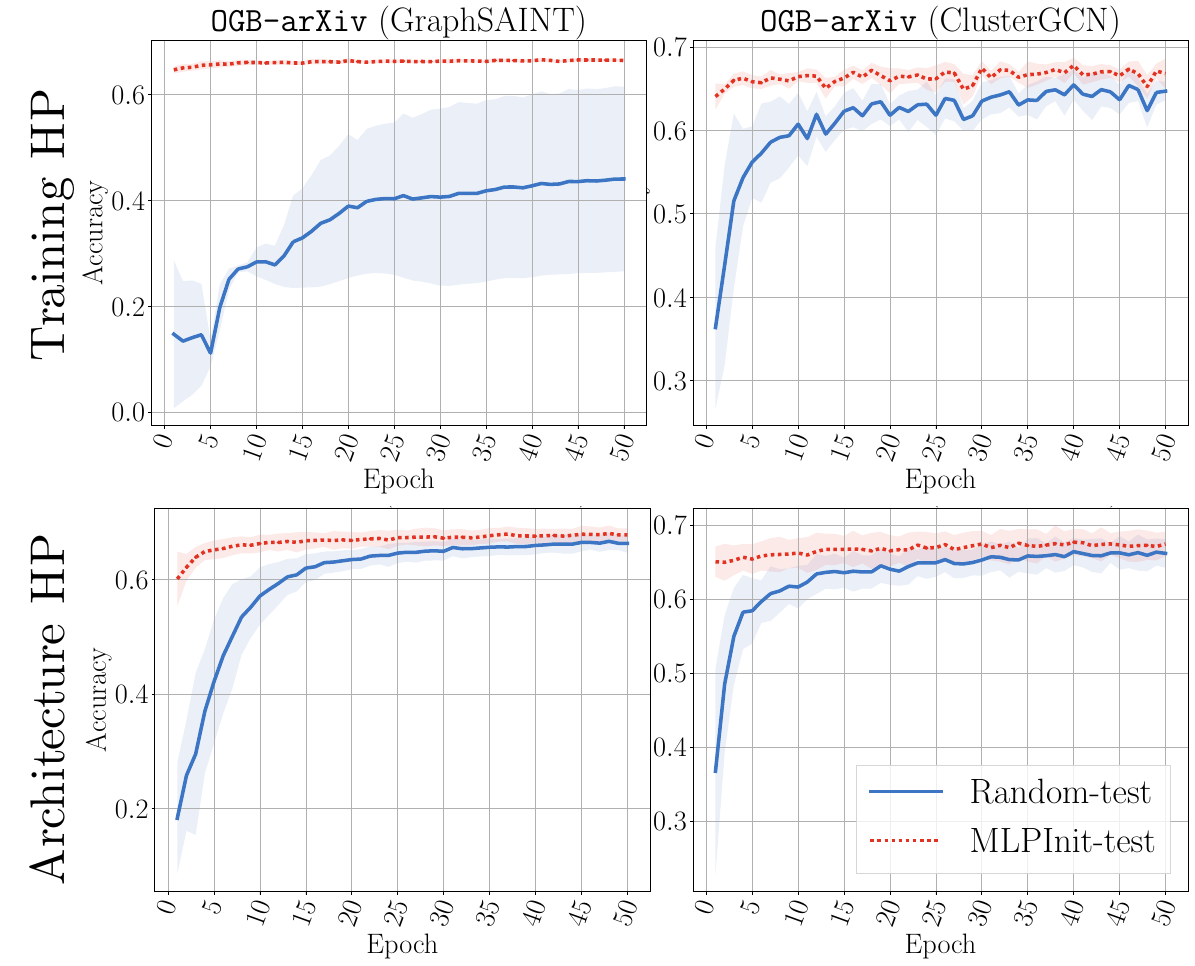}
    \vspace{-20pt}
    \caption{Training curves of GNNs with various hyperparameters. GNNs with \mlpinit consistently outperform random initialization and have a smaller standard deviation.}\label{fig:ogbn_arxiv_hps}
\end{wrapfigure}

\textbf{Observation 5: {\mlpinit}  makes GNNs less sensitive to hyperparameters and improves the overall performance across various hyperparameters.} 
In this experiment, we trained {\peermlp} and GNN with different ``Training HP'' (Learning rate, weight decay, and batch size) and ``Architecture HP'' (i.e., layers, number of hidden neurons), and we presented the learning curves of GNN with different hyperparameters in \cref{fig:ogbn_arxiv_hps}. One can see from the results that GNNs trained with \mlpinit have a much smaller standard deviation than those trained with random initialization. Moreover, \mlpinit consistently outperforms random initialization in task performance. This advantage allows our approach to saving time in searching for architectural hyperparameters. In practice, different datasets require different hyperparameters. Using our proposed method, we can generally choose random hyperparameters and obtain reasonable and relatively stable performance owing to the {\peermlp}'s lower sensitivity to hyperparameters.

\begin{figure}[t]
    \centering
    \begin{minipage}{0.48\textwidth}
        \centering
        \includegraphics[width=1.0\linewidth]{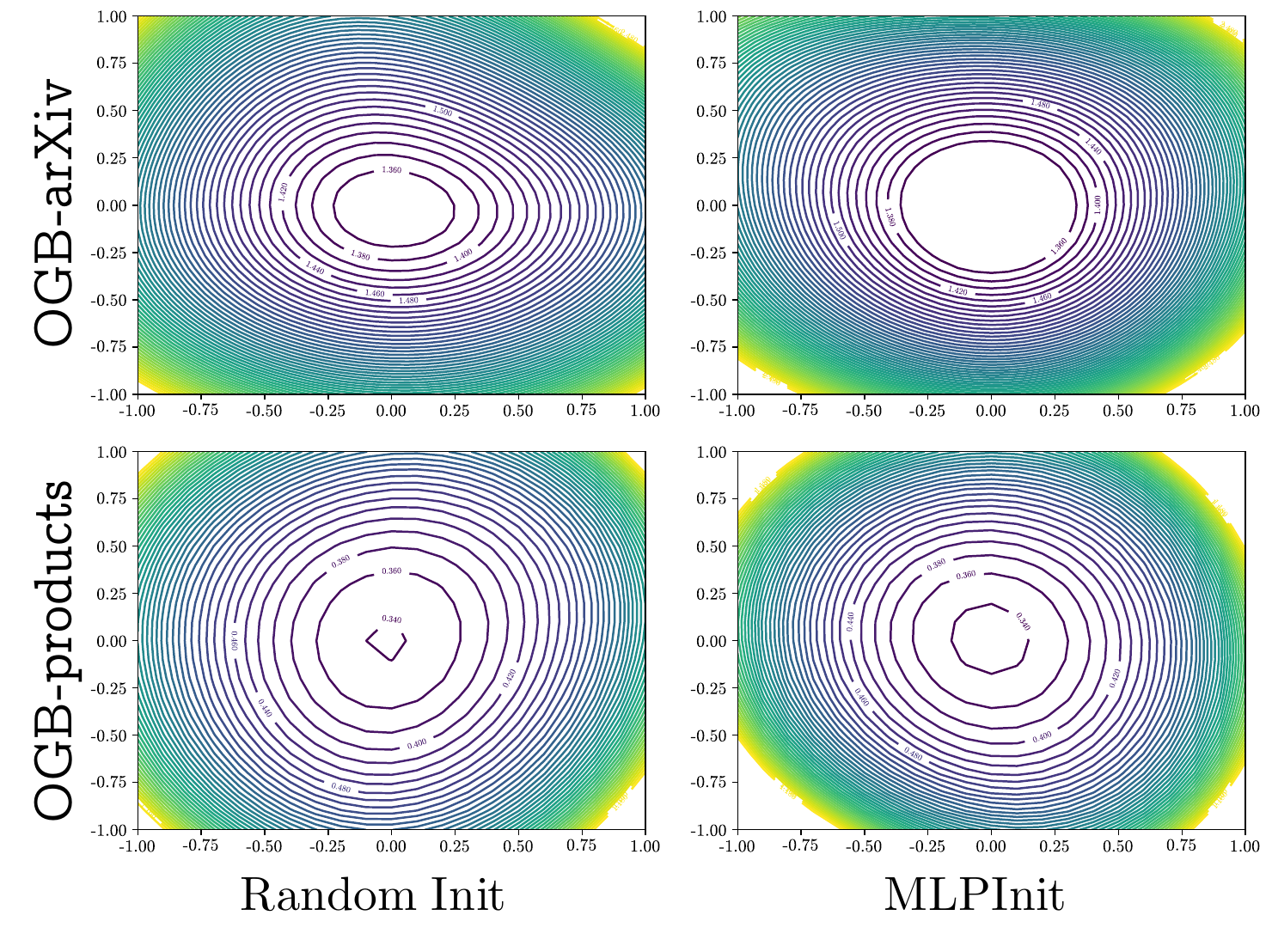}\vspace{-12pt}
        \caption{The loss landscape of GNN trained with random initialization (left) and {\mlpinit} (right). The low-loss area of GNNs with \mlpinit is larger than that with random initialization.}\label{fig:londscape}
    \end{minipage}%
    \hfill
    \begin{minipage}{0.48\textwidth}
        \centering
        \includegraphics[width=1.0\linewidth]{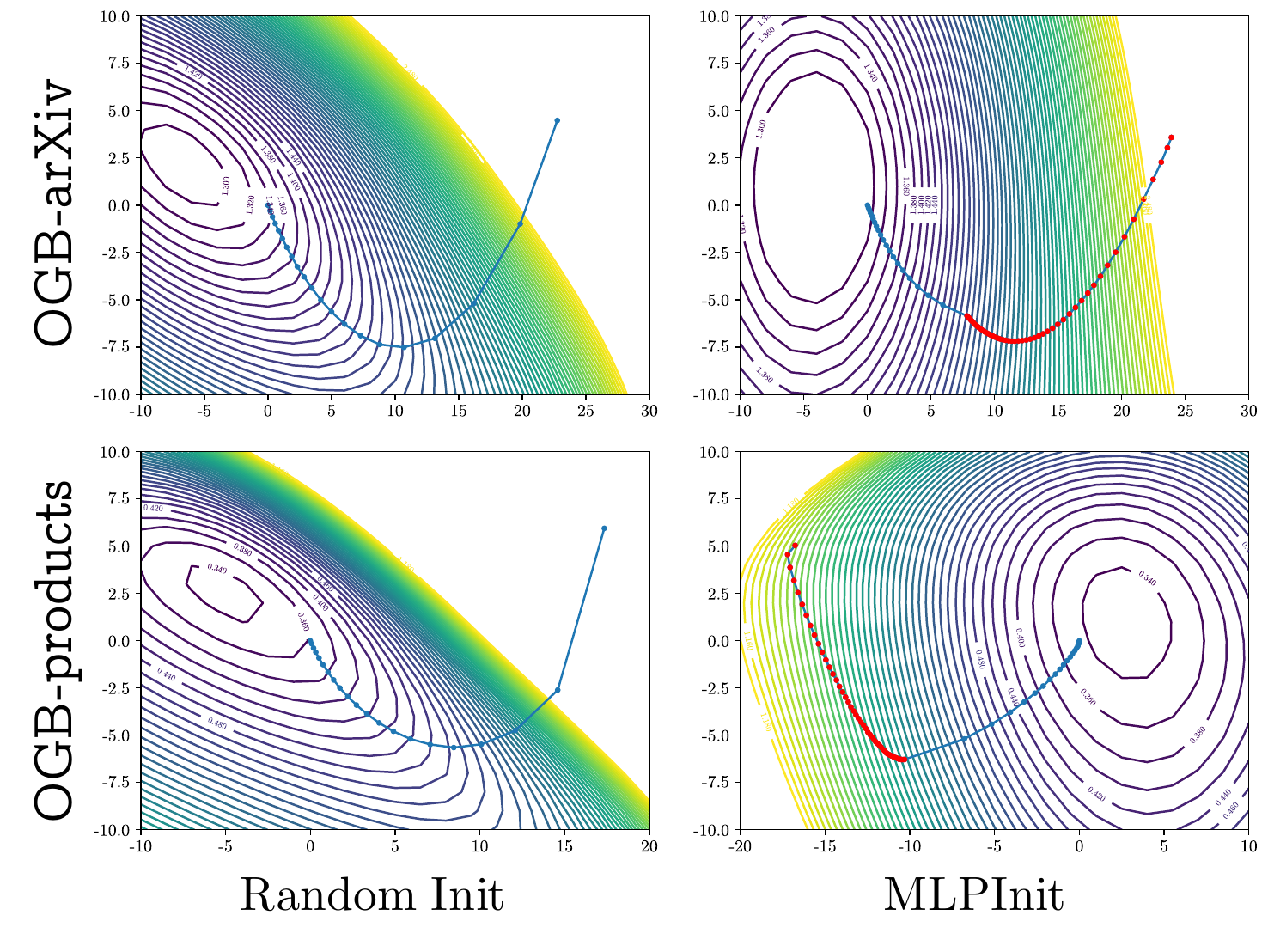}\vspace{-12pt}
        \caption{The training trajectory of the GNN with random initialization (left) and {\mlpinit} (right). The first-phase training of GNNs can be taken over by lightweight MLPs.}\label{fig:trajectory}
    \end{minipage}
    \vspace{-8pt}
\end{figure}

\subsection{Will {\mlpinit} facilitate better convergence for GNNs?}\label{sec:exp_betterconvergence}
In this experiment, we performed experiments to analyze the convergence of a fine-tuned GNN model. In other words, does this pre-training actually help find better local minima?

\textbf{Observation 6: {\mlpinit} finds larger low-loss area in loss landscape for GNNs.} The geometry of loss surfaces reflects the properties of the learned model, which provides various insights to assess the generalization ability~\citep{fort2019deep,huang2019understanding,wen2018smoothout} and robustness~\citep{fort2019deep,liu2020loss,engstrom2019exploring} of a neural network. In this experiment, we plot the loss landscape of GNN (GraphSAGE) with random weight initialization and {\mlpinit} using the visualization tool introduced in \citet{li2018visualizing}. The loss landscapes are plotted based on the training loss and with \ogbnarxiv and \ogbnproducts datasets. The loss landscapes are shown in \cref{fig:londscape}. For the fair comparison, two random directions of the loss landscape are the same and the lowest values of losses in the landscape are the same within one dataset, $1.360$ for \ogbnarxiv and $0.340$ for \ogbnproducts. From the loss landscape, we can see that the low-loss area of \mlpinit is larger in terms of the same level of loss in each dataset's (row’s) plots, indicating the loss landscape of the model trained with {\mlpinit} results in larger low-loss area than the model with random initialization. In summary, {\mlpinit} helps larger low-loss areas in loss landscape for GNNs.

\textbf{Observation 7: {\mlpinit} speeds up the optimization process} for GNNs. To further understand the training procedure of \mlpinit, we visualize the training trajectories along with loss landscapes through tools~\citep{li2018visualizing} in \cref{fig:trajectory}. In this experiment, we use GraphSAGE on \ogbnarxiv and \ogbnproducts datasets, we first train a \peermlp and then use the weights of the converged \peermlp to initialize the GraphSAGE and fine-tune it. Then we plot the training trajectories of \peermlp and GraphSAGE on the loss landscape of GraphSAGE. The {\color{myred}red} line indicates the training trajectories of the training of \peermlp and the {\color{blue}blue} line indicates the fine-tuning of GNNs. We can see that the end point of the training of MLP (red line) is close to the minima area in the loss landscape. The training trajectory clearly shows the reason why \mlpinit works, i.e., the first-phase training of GNNs can be taken over by lightweight MLPs.

\section{Related Work}\label{sec:rela}

In this section, we present several lines of related work and discuss their difference and relation to our work. Also appearing at the same conference as this work, \citep{yang2022graph} concurrently found a similar phonnomenia as our findings and provided a theoretical analysis of it.

\textbf{Message passing-based GNNs.} 
Graph neural networks typically follow the message passing mechanism, which aggregates the information from node's neighbors and learns node representation for downstream tasks. Following the pioneering work GCN~\citep{kipf2016semi}, several other works~\citep{velivckovic2018graph,xu2018powerful,balcilar2021analyzing,thorpe2021grand++,brody2021attentive,tailor2021we} seek to improve or better understand message passing-based GNNs.

\textbf{Efficient GNNs.} 
In order to scale GNNs to large-scale graphs, the efficiency of GNNs has attracted considerable recent attention. Subgraph sampling technique \citep{hamilton2017inductive,chiang2019cluster,zeng2019graphsaint} has been proposed for efficient mini-batch training for large-scale graphs. All of these methods follow the message passing mechanism for efficient GNN training. There is another line of work on scalable GNNs, which uses MLPs to simplify GNNs~\citep{zhang2022graph,wu2019simplifying,frasca2020sign,sun2021scalable,huang2020combining,hu2021graph}. These methods aim to decouple the feature aggregation and transformation operations to avoid excessive, expensive aggregation operations. We compared our work with this line of work in \cref{sec:meth:dissuss}. There are also other acceleration methods, which leverage weight quantization and graph sparsification~\citep{cai2020graph}. However, these kinds of methods often sacrifice prediction accuracy and increase modeling complexity, while sometimes meriting significant additional engineering efforts.

\textbf{GNN Pre-training.} The recent GNN pretraining methods mainly adopt contrastive learning \citep{hassani2020contrastive,qiu2020gcc,zhu2020deep,zhu2021graph, you2021graph,you2020does,jin2020self,han2022geometric,zhu2020self}. GNN pretraining methods typically leverage graph augmentation to pretrain weights of GNNs or obtain the node representation for downstream tasks. For example, \cite{zhu2020deep} maximizes the agreement between two views of one graph. GNN pre-training methods not only use graph information for model training but also involve extra graph data augmentation operations, which require additional training time and engineering effort.

\section{Conclusion}\label{sec:concl}
This work presents a simple yet effective initialization method, {\mlpinit}, to accelerate the training of GNNs, which adopts the weights from their converged \peermlp initialize GNN and then fine-tune GNNs. With comprehensive experimental evidence, we demonstrate the superiority of our proposed method on training speedup (up to 33× speedup on \ogbnproducts), downstream task performance improvements(up to $7.97\%$ performance improvement for GraphSAGE), and robustness improvements (larger minimal area in loss landscape) on the resulting GNNs. Notably, our proposed method is easy to implement and employ in real applications to speed up the training of GNNs.

\clearpage
\section*{Acknowledgement}
We thank the anonymous reviewers for their constructive suggestions and fruitful discussion. Xiaotian would like to thank Zirui Liu from Rice University for the discussion about the training time for MLP and GNN. He would also like to thank Kaixiong Zhou from Rice University and Keyu Duan from National University of Singapore for the discussion about the GNN training on large-scale graphs. Xiaotian would also like to thank Hanqing Zen from Meta Inc., for his valuable feedback and suggestions on the manuscript of this work. We thank Jingyuan Li from the Department of Electrical and Computer Engineering at the University of Washington for identifying a typo in our paper. Portions of this research were conducted with the advanced computing resources provided by Texas A\&M High Performance Research Computing.  This work is, in part, supported by NSF IIS-1750074 and IIS-1900990. The views and conclusions contained in this paper are those of the authors and should not be interpreted as representing any funding agencies.

\section*{Reproducibility Statement}
To ensure the reproducibility of our experiments and benefit the research community, we provide the source code at \href{https://github.com/snap-research/MLPInit-for-GNNs}{\color{blue} {https://github.com/snap-research/MLPInit-for-GNNs}}. The hyper-parameters and other variables required to reproduce our experiments are described in \cref{sec:app:imple}.

\bibliography{ref}
\bibliographystyle{iclr2023_conference}

\clearpage
\appendix

\part{}
\localtableofcontents
\clearpage

\section{Additional Experiments}\label{sec:app:add}
In this appendix, we present additional experiments to show the superiority of our proposed method, \mlpinit, including additional results for link prediction and training curves. We also present more analysis on the weights distribution changes of \mlpinit.

\subsection{Additional experimental link prediction}\label{sec:app:add:lp}
Here, we present the additional experiment results on the link prediction task in \cref{tab:lp_app}, which is similar to our results in \cref{tab:lp}, but on more datasets. In general, we observe that GNN with \mlpinit outperforms that with random initialization on the link prediction tasks on these additional datasets. \mlpinit does not gain better performance on \aproducts dataset. We conjecture that the reason for this is that node features may contain less task-relevant information on \aproducts.

\begin{table}[!htb]
\centering
\fontsize{10}{12}\selectfont
\setlength{\tabcolsep}{6pt}
\caption{The performance of link prediction task. The results are based on ten runs.}\label{tab:lp_app}
\vspace{-5pt}
\begin{tabular}{l|l|ccrrrrrrr} 
\toprule
                            &Methods                                &AUC                &AP                 & Hits@10           & Hits@20           & Hits@50            & Hits@100 \\ \midrule
\multirow{4}{*}{\rotatebox{90}{\scriptsize \cora}}       & $\textrm{MLP}_{\textrm{random}}$      &\ms{91.87} {1.25}	&\ms{92.16} {0.99}	&\ms{46.98} {4.33}	&\ms{58.05} {6.44}	&\ms{76.03} {4.17}	&\ms{86.28} {3.05} \\
                            & $\textrm{GNN}_{\textrm{random}}$      &\ms{91.80} {1.39}	&\ms{92.68} {1.28}	&\ms{51.48} {7.57}	&\ms{63.04} {6.12}	&\ms{78.12} {4.06}	&\ms{86.07} {2.77} \\
                            & $\textrm{GNN}_{\textrm{mlpinit}}$     &\ms{92.93} {0.88}	&\ms{93.36} {0.86}	&\ms{53.72} {7.43}	&\ms{68.65} {4.22}	&\ms{80.30} {2.35}	&\ms{89.37} {2.21} \\
                            & Improvement      &{\color{blue}$\uparrow 1.23\%$}	&{\color{blue}$\uparrow 0.74\%$}	&{\color{blue}$\uparrow 4.35\%$}	&{\color{blue}$\uparrow 8.91\%$}	&{\color{blue}$\uparrow 2.79\%$}	&{\color{blue}$\uparrow 3.84\%$} \\ \midrule

\multirow{4}{*}{\rotatebox{90}{\scriptsize\seer }}   & $\textrm{MLP}_{\textrm{random}}$      &\ms{90.10} {0.99}	&\ms{90.65} {0.98}	&\ms{46.64} {5.11}	&\ms{57.21} {4.84}	&\ms{71.45} {3.15}	&\ms{81.93} {2.24} \\
                            & $\textrm{GNN}_{\textrm{random}}$      &\ms{89.86} {1.18}	&\ms{90.88} {0.99}	&\ms{49.98} {5.54}	&\ms{58.02} {5.00}	&\ms{71.65} {3.50}	&\ms{82.11} {2.84} \\
                            & $\textrm{GNN}_{\textrm{mlpinit}}$     &\ms{90.51} {1.07}	&\ms{90.96} {0.80}	&\ms{50.02} {3.09}	&\ms{58.77} {4.48}	&\ms{71.78} {4.12}	&\ms{82.66} {2.53} \\
                            & Improvement   &{\color{blue}$\uparrow 0.73\%$}	&{\color{blue}$\uparrow 0.08\%$}	&{\color{blue}$\uparrow 0.09\%$}	&{\color{blue}$\uparrow 1.29\%$}	&{\color{blue}$\uparrow 0.18\%$}	&{\color{blue}$\uparrow 0.67\%$}      \\ \midrule
       
\multirow{4}{*}{\rotatebox{90}{\scriptsize\cs}}         & $\textrm{MLP}_{\textrm{random}}$      &\ms{96.29} {0.12}	&\ms{95.79} {0.13}	&\ms{13.36} {1.49}	&\ms{19.67} {2.21}	&\ms{33.46} {2.17}	&\ms{46.82} {1.91} \\
                            & $\textrm{GNN}_{\textrm{random}}$      &\ms{96.11} {0.08}	&\ms{95.75} {0.10}	&\ms{14.27} {2.77}	&\ms{22.57} {2.52}	&\ms{35.40} {2.01}	&\ms{48.21} {2.00} \\
                            & $\textrm{GNN}_{\textrm{mlpinit}}$     &\ms{96.72} {0.10}	&\ms{96.49} {0.14}	&\ms{16.96} {3.37}	&\ms{25.44} {3.00}	&\ms{40.69} {2.99}	&\ms{53.78} {2.00} \\
                            & Improvement     &{\color{blue}$\uparrow 0.63\%$}	&{\color{blue}$\uparrow 0.77\%$}	&{\color{blue}$\uparrow 18.81\%$}	&{\color{blue}$\uparrow 12.70\%$}	&{\color{blue}$\uparrow 14.96\%$}	&{\color{blue}$\uparrow 11.55\%$}                \\ \midrule

\multirow{4}{*}{\rotatebox{90}{\scriptsize\computers}}  & $\textrm{MLP}_{\textrm{random}}$      &\ms{81.85} {0.79}	&\ms{82.41} {0.69}	&\ms{2.10} {0.48}	&\ms{4.13} {0.86}	&\ms{7.83} {0.95}	&\ms{12.18} {1.01} \\
                            & $\textrm{GNN}_{\textrm{random}}$      &\ms{91.78} {0.48}	&\ms{91.94} {0.42}	&\ms{7.60} {1.47}	&\ms{11.10} {1.74}	&\ms{18.64} {1.94}	&\ms{25.42} {2.15} \\
                            & $\textrm{GNN}_{\textrm{mlpinit}}$     &\ms{90.76} {1.61}	&\ms{91.06} {1.47}	&\ms{6.76} {3.27}	&\ms{11.11} {1.82}	&\ms{17.40} {2.58}	&\ms{24.59} {2.56} \\
                            & Improvement      &{\color{myred}$\downarrow 1.11\%$}	&{\color{myred}$\downarrow 0.96\%$}	&{\color{myred}$\downarrow 11.04\%$}	&{\color{blue}$\uparrow 0.16\%$}	&{\color{myred}$\downarrow 6.65\%$}	&{\color{myred}$\downarrow 3.26\%$}                \\ \midrule
                            
\multirow{4}{*}{\rotatebox{90}{\scriptsize\corafull}}   & $\textrm{MLP}_{\textrm{random}}$      &\ms{95.72} {0.18}	&\ms{95.55} {0.23}	&\ms{19.38} {4.71}	&\ms{27.83} {3.27}	&\ms{42.98} {2.01}	&\ms{57.20} {1.27} \\
                            & $\textrm{GNN}_{\textrm{random}}$      &\ms{95.87} {0.36}	&\ms{95.77} {0.42}	&\ms{21.33} {4.77}	&\ms{30.57} {3.49}	&\ms{45.08} {3.46}	&\ms{59.58} {2.53} \\
                            & $\textrm{GNN}_{\textrm{mlpinit}}$     &\ms{96.71} {0.16}	&\ms{96.73} {0.22}	&\ms{25.78} {4.92}	&\ms{36.68} {5.36}	&\ms{53.81} {2.34}	&\ms{66.73} {1.96} \\
                            & Improvement      &{\color{blue}$\uparrow 0.87\%$}	&{\color{blue}$\uparrow 1.01\%$}	&{\color{blue}$\uparrow 20.87\%$}	&{\color{blue}$\uparrow 19.98\%$}	&{\color{blue}$\uparrow 19.37\%$}	&{\color{blue}$\uparrow 12.01\%$}           \\

\bottomrule
\end{tabular}
\end{table}

\subsection{Additional hyperparameter sensitivity}\label{sec:app:add:hps}
\begin{figure}[t]
    \centering
    \includegraphics[width=0.99\textwidth]{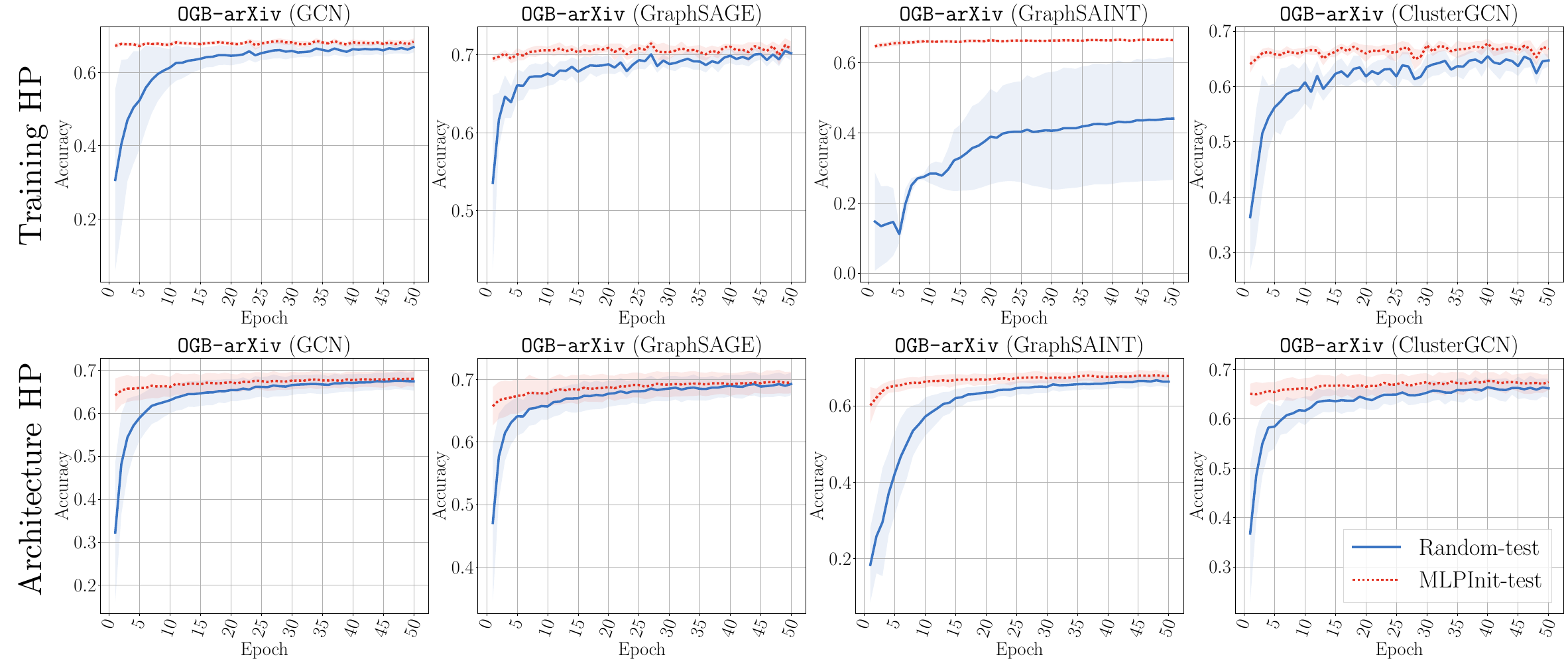}
    \caption{The training curves of the GNNs with different hyperparameters on \ogbnarxiv dataset. The training curves of GNN with \mlpinit generally obtain lower loss and higher accuracy than those with the random initialization and converge faster. The training curves are depicted based on ten runs.}\label{fig:ogbn_arxiv_hps_add}
\end{figure}

In this appendix, we present the additional results to explore the sensitivity to the various hyperparameters. The results are the full version of \cref{fig:ogbn_arxiv_hps_add}. 

\subsection{Additional training curves}

\begin{figure}[t]
     \centering
     \begin{subfigure}[b]{1.0\textwidth}
         \centering
         \includegraphics[width=\textwidth]{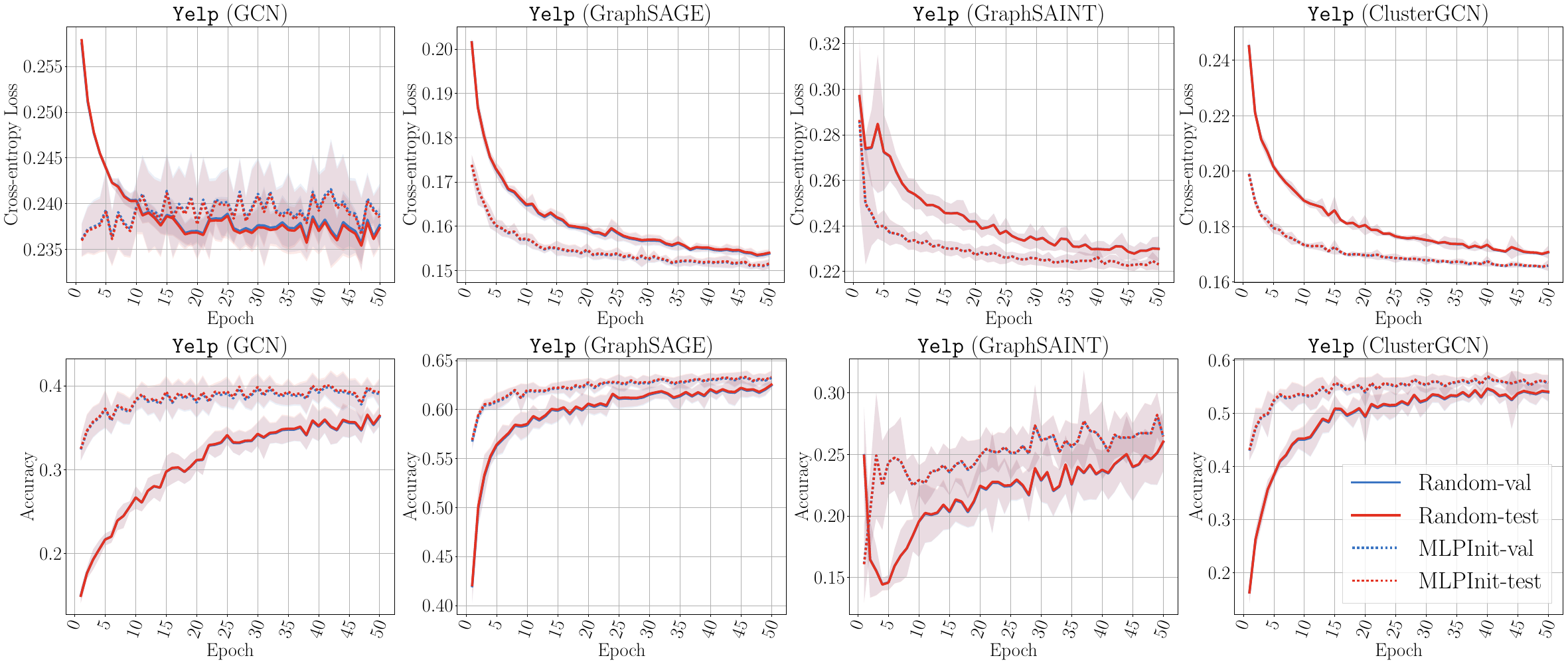}
     \end{subfigure}\\
     \begin{subfigure}[b]{1.0\textwidth}
         \centering
         \includegraphics[width=\textwidth]{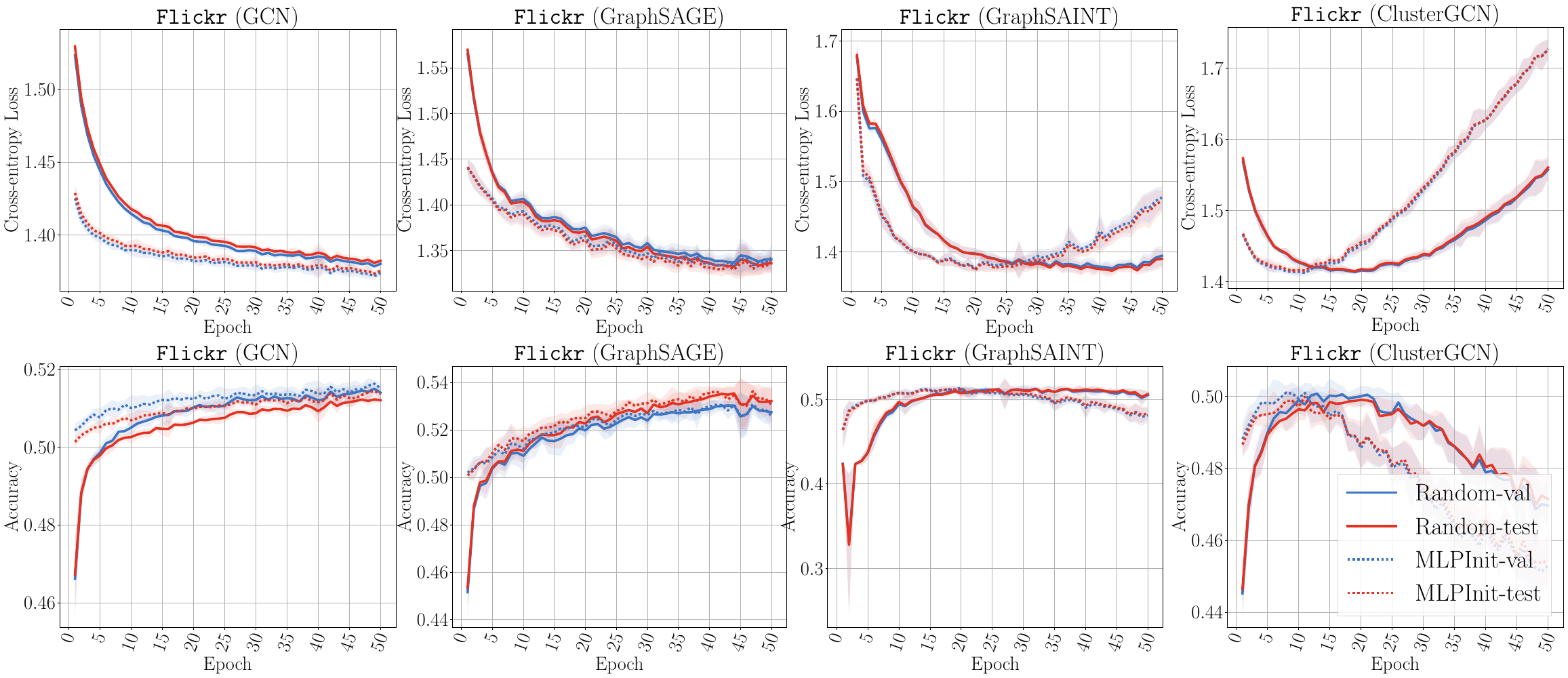}
     \end{subfigure}\\
     \begin{subfigure}[b]{1.0\textwidth}
         \centering
         \includegraphics[width=\textwidth]{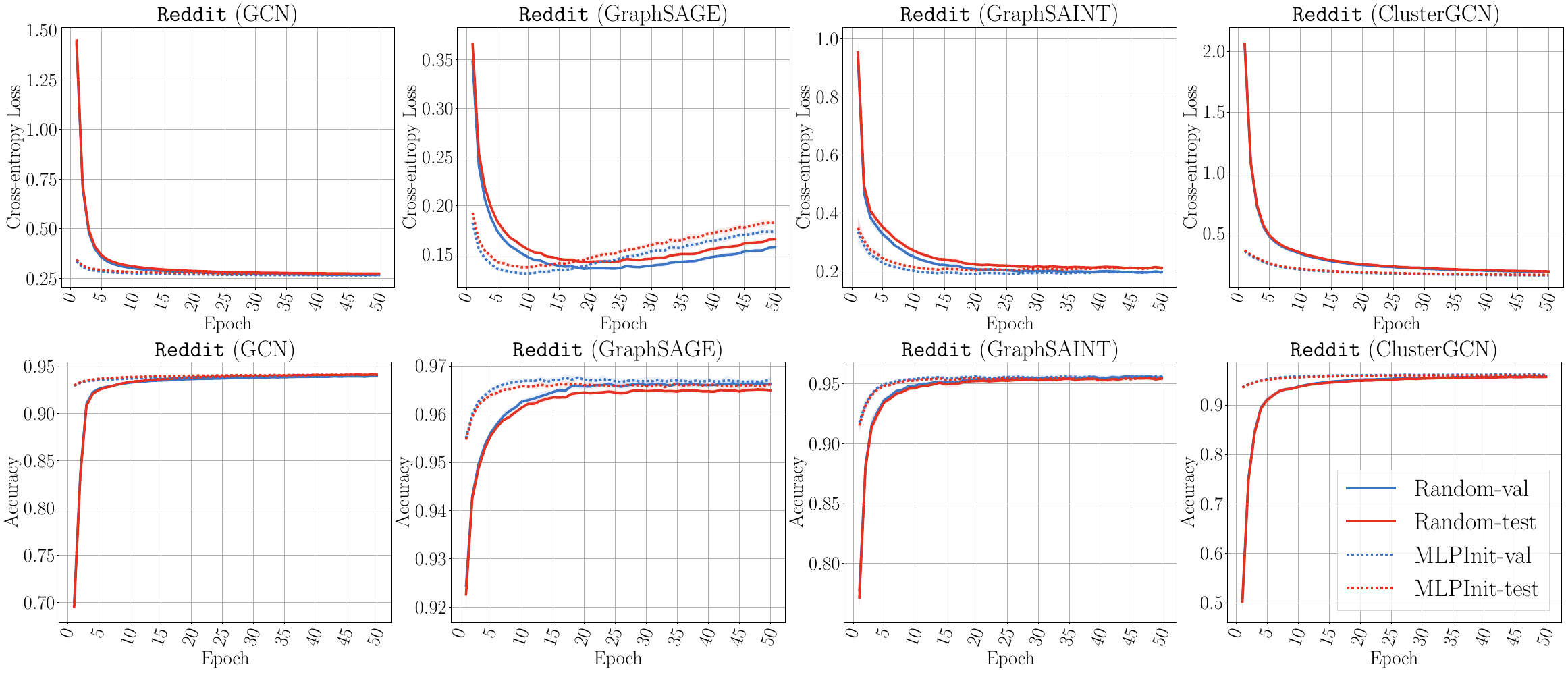}
     \end{subfigure}
\caption{The training curves of GNN with random initialization and \mlpinit on \yelp, \flickr, \reddit datasets.}\label{fig:perf_add1}
\end{figure}

\begin{figure}[t]
     \centering
     \begin{subfigure}[b]{1.0\textwidth}
         \centering
         \includegraphics[width=\textwidth]{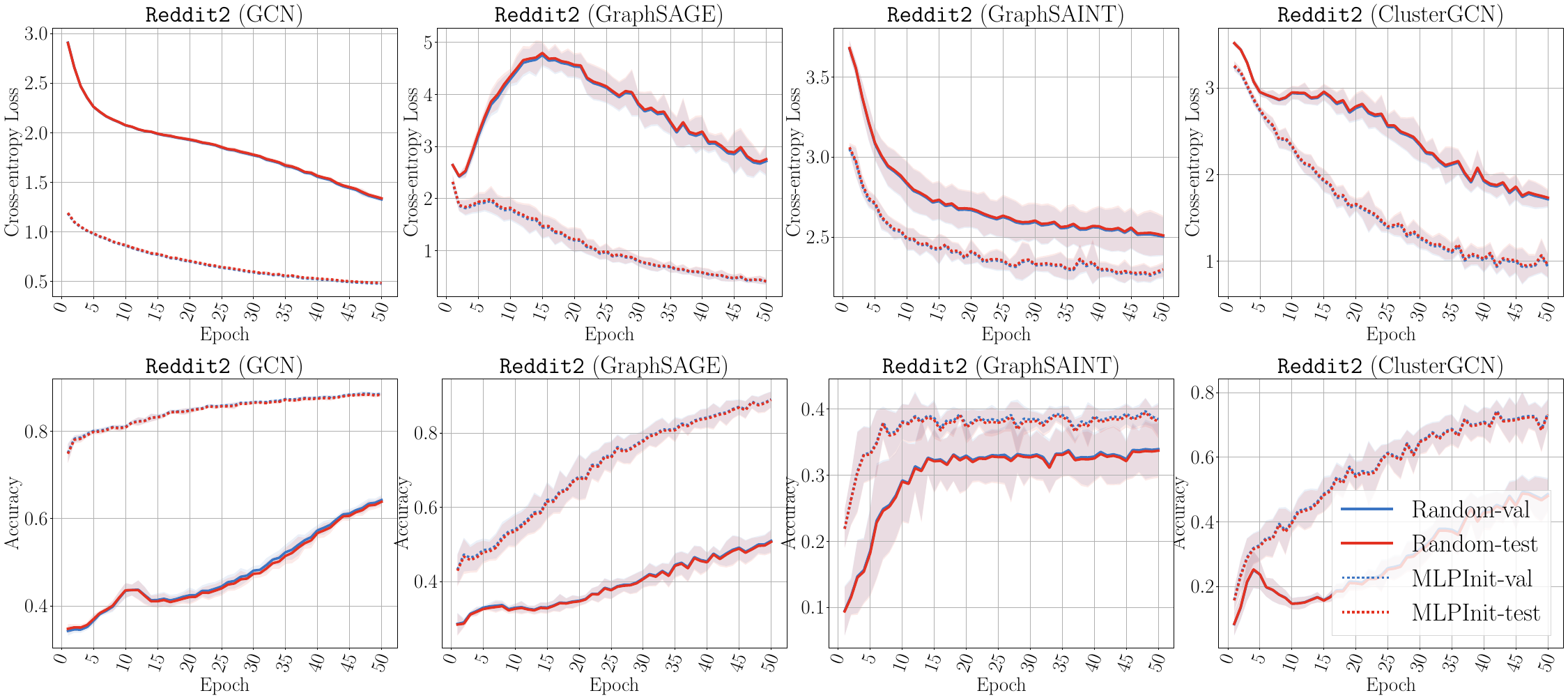}
     \end{subfigure}\\
     \begin{subfigure}[b]{1.0\textwidth}
         \centering
         \includegraphics[width=\textwidth]{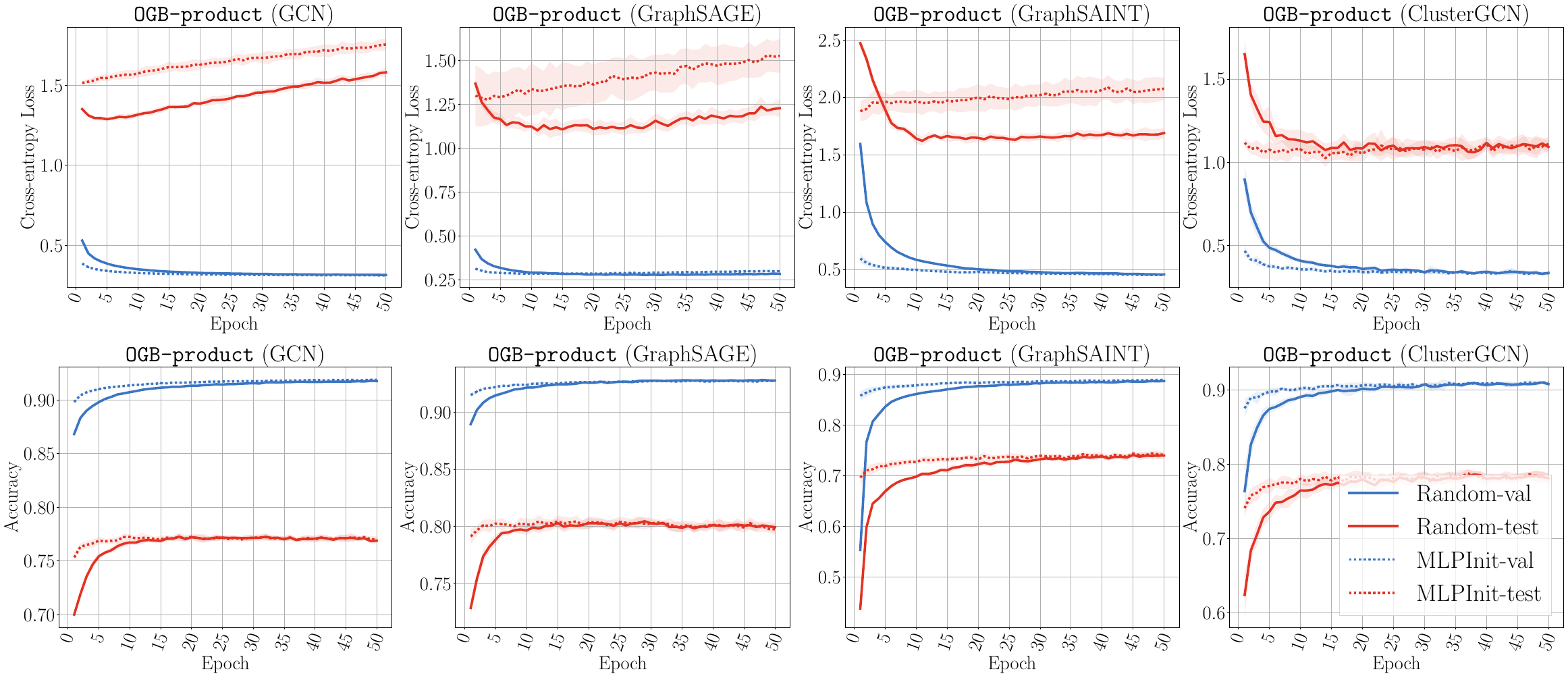}
     \end{subfigure}\\
     \begin{subfigure}[b]{1.0\textwidth}
         \centering
         \includegraphics[width=\textwidth]{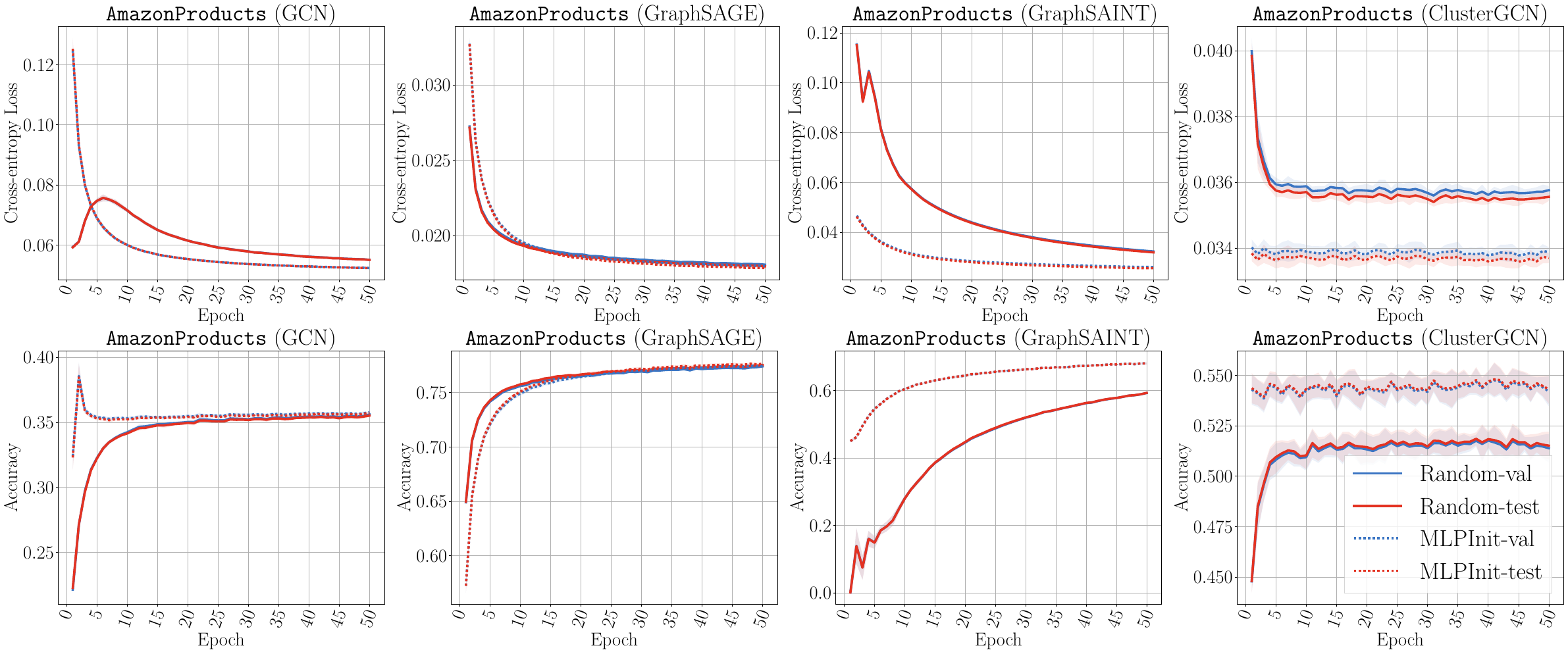}
     \end{subfigure}
\caption{The training curves of GNN with random initialization and with \mlpinit on \reddittwo, \ogbnproducts, \aproducts datasets.}\label{fig:perf_add2}
\end{figure}

In this appendix, we present the additional training curves of other datasets in \cref{fig:perf_add1} and \cref{fig:perf_add2}, which are additional experimental results of \cref{fig:ogbn_arxiv_perf}. The results comprehensively show the training curves on various datasets. As we can see from  \cref{fig:perf_add1} and \cref{fig:perf_add2} that \mlpinit consistently outperforms the random initialization and is able to accelerate the training of GNNs.

\subsection{Additional loss/accuracy curves of \peermlp and GNN }\label{sec:app:addition:loss_acc}
\begin{figure}[t]
    \centering
    \includegraphics[width=0.9\textwidth]{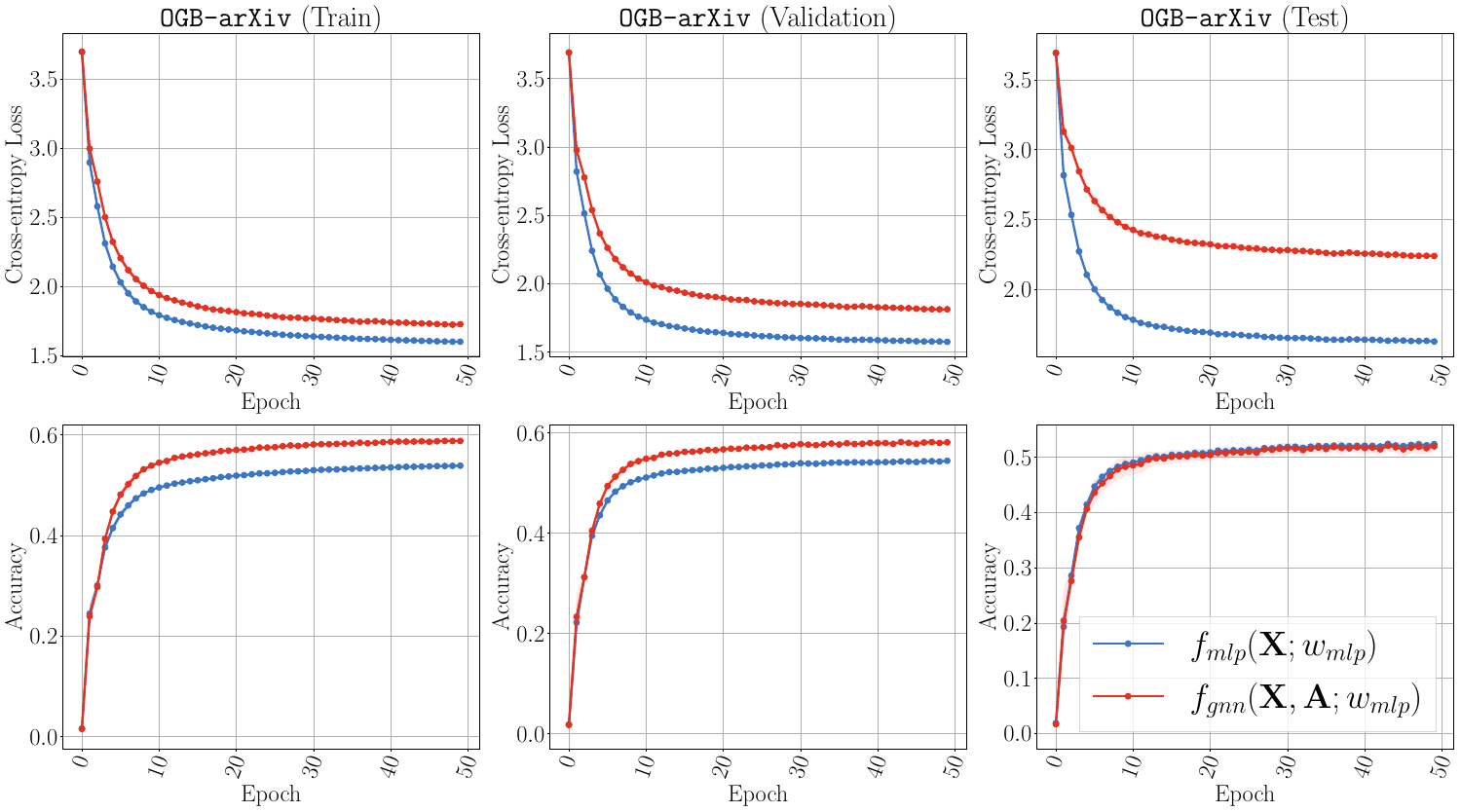}
    \caption{The relation of GNN and MLP during the training of PeerMLP. We report the cross-entropy loss and accuracy on the training/validation/test sets.}\label{fig:gnn_mlp_loss_all}
\end{figure}

In this appendix, we plotted the loss and accuracy curves of \peermlp and GNN on training/validation/test set and presented the results in \cref{fig:gnn_mlp_loss_all}, which are the additional experimental results to \cref{fig:gnn_mlp_loss}. The results surprisingly show that GNN using the weight from trained \peermlp has worse cross-entropy loss but better prediction accuracy than \peermlp. The reason would be that GNN can smooth the prediction logit, making loss worse but accuracy better.

\subsection{Training curves of link prediction task}\label{sec:app:addition:lp_curves}

To further investigate the training process of the link prediction task, we present the training curves for the link prediction task for each metric we used. The metrics we used are AUC, AP, Hits@K, which are commonly used to evaluate the performance of link prediction \citep{zhang2018link,zhang2021labeling,zhao2022learning}. AUC and AP measure binary classification, reflecting the value of loss for link prediction, and Hits@K is the count of how many positive samples are ranked in the top-K positions against a bunch of negative samples. 

The results show that the AUC and AP are easy to train, while the Hits@K is harder to train. The GNN with random initialization needs much more time than \mlpinit to obtain a good Hits@K. Since Hits@K is a more realistic metric, the results demonstrate the superiority of our method in the link prediction task. These experimental results also show that node features are important for link prediction task since we can obtain a good performance only with MLP, therefore, \mlpinit is beneficial for link prediction since the \mlpinit used node feature information only to train \peermlp.

\begin{figure}[t]
    \centering
    \includegraphics[width=0.95\textwidth]{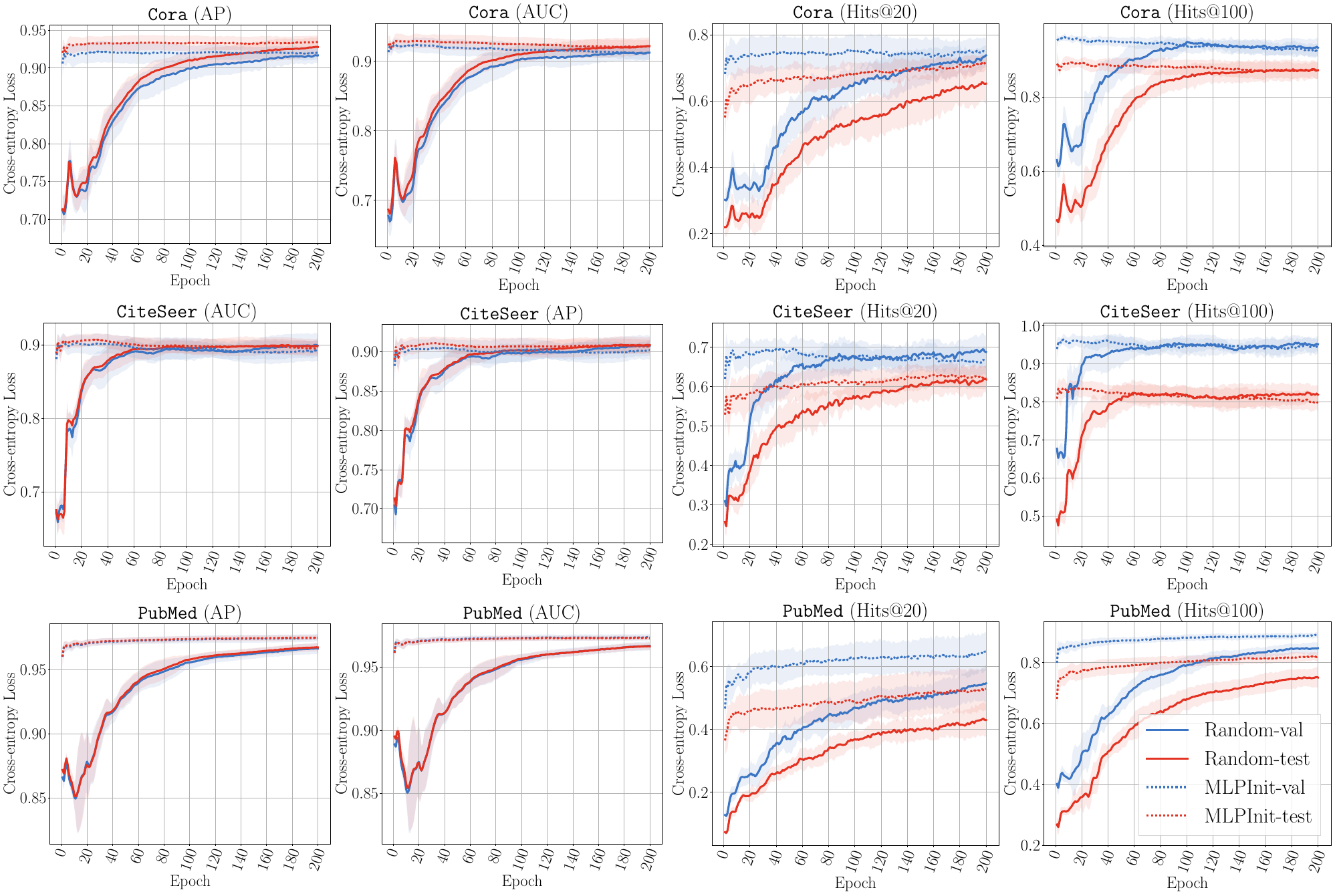}
    \caption{The training curves for link prediction task on \cora, \seer, \pubmed datasets. GNN with \mlpinit generally obtains higher metrics for link prediction task than those with random initialization and converge faster.}\label{fig:lp_curves}
\end{figure}

\clearpage
\section{More Experiments}

In this appendix, we conducted additional experiments to further analyze our proposed method \mlpinit. We conducted experiments to show that training MLP is much cheaper than training GNN. The results show that the running time of MLP can be negligible compared to that of GNNs. We compare the efficiency of our method to GNN pre-training methods. The comparison to GNN pre-training methods demonstrates the superiority of \mlpinit in effectiveness and efficiency. We conduct experiments to investigate the effectiveness of \mlpinit on GNN with more complicated aggregators. We provided the new form of our results in \cref{tab:perf} to show the performance improvements for graph sampling methods and GNN architectures separately.

\subsection{The Performance of GCN with the weight of \peermlp }\label{sec:app:add:gcn_cora}
In this experiment, we aimed to verify \textit{Observation 2: Converged weights from {\peermlp} provide a good GNN initialization} from \cref{sec:what_will_happen} by evaluating the performance of GCN with the weight of \peermlp on the \cora, \seer, and \pubmed datasets. We trained the \peermlp for GCN, and then calculated the accuracy of the GCN using the well-trained \peermlp weights (without fine-tuning). We used the public split for these three datasets in a semi-supervised setting. The evaluated models are described in detail below and the results are presented in \cref{tab:gnn_mlp_cora}.
\begin{itemize}
    \item {\textsf{PeerMLP}}: the well-trained \peermlp for GCN on different datasets.
    \item GCN w/ $w_{\textsf{peermlp}}$: the GCN with the weight of well-trained \peermlp, without fine-tuning.
    \item GCN: the well-trained graph convolutional neural network.
\end{itemize}

\begin{table}[h]
\centering
\fontsize{9}{12}\selectfont
\caption{The performance of GNNs and its {\peermlp} with the weights of a converged {\peermlp} on test data. The accuracy are in percentage ($\%$).}\label{tab:gnn_mlp_cora}
\vspace{-8pt}
\begin{tabular}{l|ccc|c} 
\toprule
                                                  & {\textsf{PeerMLP}}   & GCN w/ $w_{\textsf{peermlp}}$   & Improv.                           &GCN    \\ \midrule
\multirow{1}{*}{\rotatebox{0}{{\cora}}}           & 58.50                &77.60                            &{\color{blue}$\uparrow 32.64\%$}  &82.60  \\
\multirow{1}{*}{\rotatebox{0}{{\seer}}}           & 60.50                &69.70                            &{\color{blue}$\uparrow 15.20\%$}  &71.60  \\
\multirow{1}{*}{\rotatebox{0}{{\pubmed}}}         & 73.60                &78.10                            &{\color{blue}$\uparrow 6.11\%$}  &79.80  \\
\bottomrule
\end{tabular}
\end{table}

The results show  that the GNN w/ $w_{\textsf{peermlp}}$ significantly outperforms the \peermlp, even though the weight $w_{\textsf{peermlp}}$ is trained on \peermlp. The performance improvements are notable, with increases of $32.64\%$, $15.20\%$, and $6.11\%$ on \cora, \seer, and \pubmed datasets, respectively. The results would be additional evidence for \textit{Observation 2: Converged weights from {\peermlp} provide a good GNN initialization.}

Moreover, this intriguing phenomenon implies a relationship between MLPs and GNNs that could potentially shed light on the generalization capabilities of GNNs. We believe that our findings will be of significant interest to researchers and practitioners in the field of graph neural networks, and we hope that our work will inspire follow-up research to further explore the relationship between MLPs and GNNs.

\subsection{Weight difference of GNNs with random initialization and \mlpinit }\label{sec:app:add:weight}
\begin{figure}[!htb]
     \centering
     \begin{subfigure}[b]{0.4\textwidth}
         \centering
         \includegraphics[width=\textwidth]{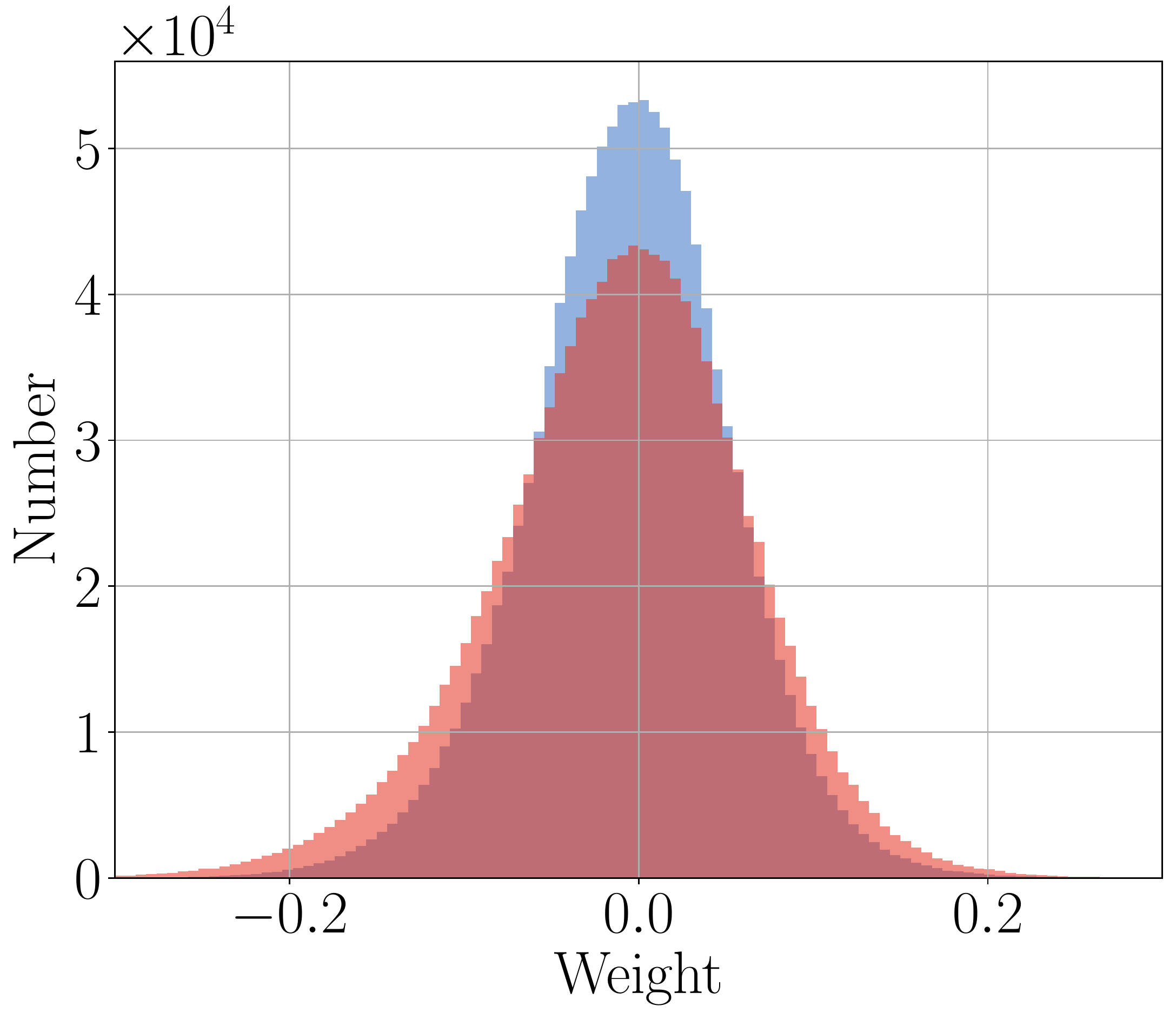}
         \caption{\texttt{OGB-arXiv}}\label{fig:weight:arxiv}
     \end{subfigure} 
     \begin{subfigure}[b]{0.4\textwidth}
         \centering
         \includegraphics[width=\textwidth]{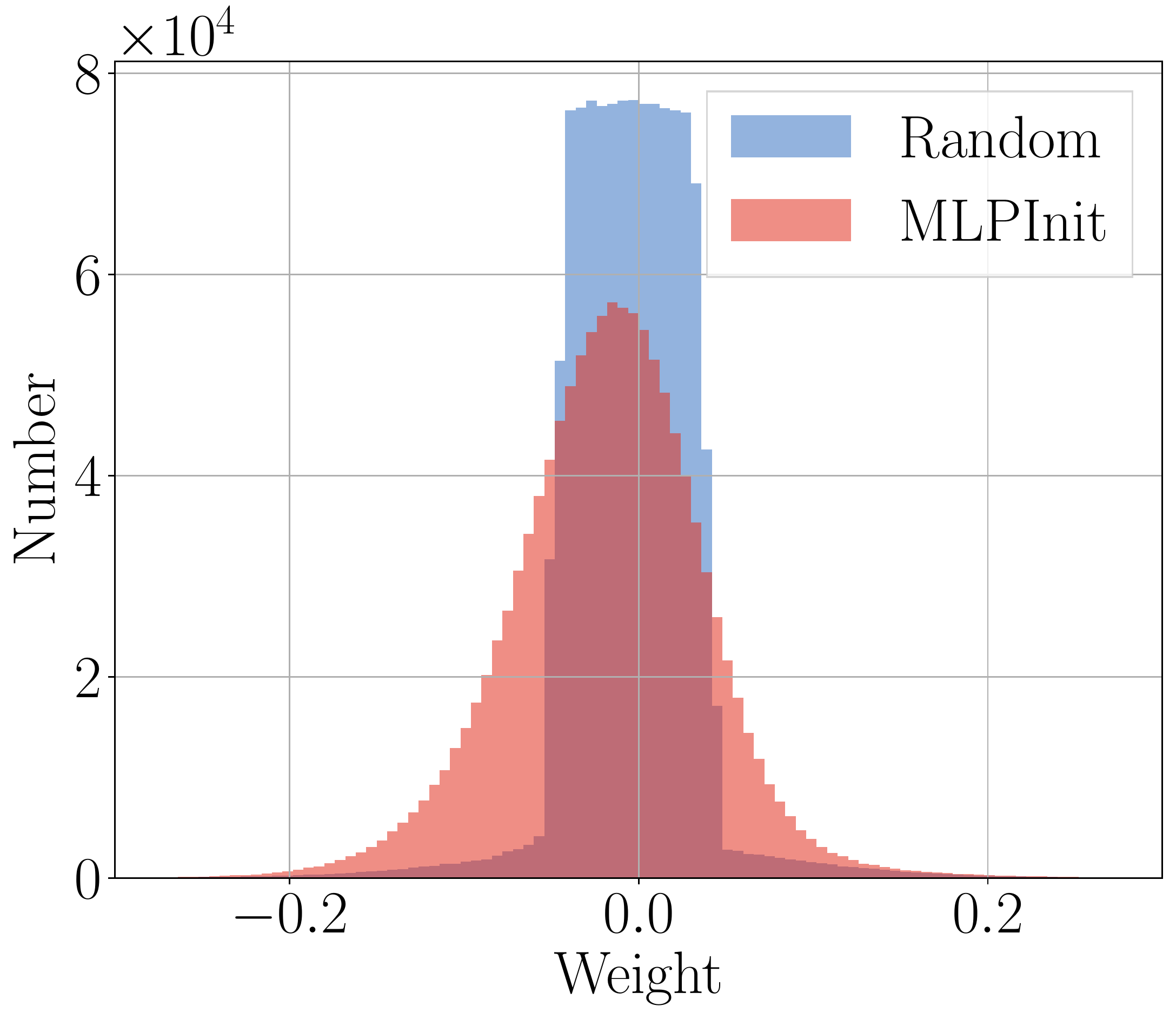}
         \caption{\texttt{OGB-products}}\label{fig:weight:products}
     \end{subfigure}
\caption{Weight histograms of GraphSAGE model weights with random initialization (blue) and \mlpinit (red) on \ogbnarxiv (left) and \ogbnproducts (right). The results are plotted based on $20$ epochs after training GraphSAGE with zero weight decay. Note that \mlpinit produces more weights of higher magnitude.}\label{fig:weight}
\end{figure}

Prior work \citep{li2018visualizing} suggests that ``small weights still appear more sensitive to perturbations, and produce sharper looking minimizers.'' To this end, we explore the distribution of weights of GNNs with both random initialization and \mlpinit, and present the results in \cref{fig:weight}. We can observe that with the same number of training epochs, the weights of GraphSAGE with \mlpinit produce more high-magnitude (both positive and negative) weights, indicating the \mlpinit can help the optimization of GNN. This difference stems from a straightforward reason: \mlpinit provides a good initialization for GNNs since the weights are trained by the \peermlp before (also aligning with our observations in \cref{sec:exp_betterconvergence}).

\subsection{Running time comparison of MLP and GNN}
We conducted experiments to compare the running time of MLP and GNN (GraphSAGE in this experiment) and presented the running time needed by training MLP and GraphSAGE for one epoch in \cref{tab:mlp_gnn_run_time}. In this experiment, MLP is trained in a full-batch way, and training node features are stored in GPU memory. We adopted the official example code of GraphSAGE \footnote{\url{https://github.com/pyg-team/pytorch_geometric/blob/2.0.4/examples/ogbn_products_sage.py}}. The running time of MLP only needs $1/147$ and $1/2303$ of that of GraphSAGE on \ogbnarxiv and \ogbnproducts datasets. The results show that training MLP is much cheaper than training GNN. In practice, MLP usually only needs to be trained less than $50$ epochs to converge. Thus, the training time of MLP in \mlpinit is negligible compared to the training time of GNNs.

\begin{table}[H]
\centering
\caption{The comparison of the running time of MLP and GraphSAGE. We report the running time of one epoch for MLP, GraphSAGE, and their ratio. The results show that training MLP is much cheaper than training GrapSAGE, especially for the large graph.}\label{tab:mlp_gnn_run_time}
\begin{tabular}{lrrb}
\toprule
Dataset             &  MLP                  & GraphSAGE                  & MLP/GraphSAGE Ratio  \\ \midrule
\ogbnarxiv          & \ms{0.035}{0.000}     & \ms{5.170}{0.313}          & $\mathbf{1/147}$            \\
\ogbnproducts       & \ms{0.076}{0.000}     & \ms{175.758}{9.560}        & $\mathbf{1/2303}$          \\
\bottomrule
\end{tabular}
\end{table}

\subsection{Comparison to GNN pre-training methods}\label{sec:app:addition:pre}

In this appendix, we compare the efficiency of our method to GNN pre-training methods. In this experiment, we adopt DGI \citep{velivckovic2018deep} as the pre-training method to pretrain the weight of GNN. DGI maximizes the mutual information between patch representations and corresponding high-level summaries of graphs. Since the output of DGI is a hidden representation, we leverage DGI to pretrain weights of the GNN except for the last layer (classification head). We report the final prediction performance of GNN with \mlpinit and DGI in \cref{tab:dgi_perf} and reported the running time of MLPInit and DGI in \cref{tab:dgi_run_time}. The experimental results show that \mlpinit obtains rank $1.28$ and $1.42$ on GraphSAGE and GCN, demonstrating \mlpinit outperforms DGI slightly. This might be because the additional classification head for DGI is not pretrained.  It is worth noting that DGI is much more time-consuming than \mlpinit, as \cref{tab:dgi_run_time} show that \mlpinit only needs $1/6.59$ and $1/1017.41$ running time of DGI. The comparison to GNN pre-training methods demonstrates the superiority of \mlpinit in effectiveness and efficiency.

\begin{table}[t]
    \fontsize{9}{10}\selectfont
    \setlength{\tabcolsep}{2.7pt}
    \centering
    \caption{The comparison of the performance of \mlpinit and DGI. The best performance is in \textbf{boldface}. The Avg.Rank is the average performance rank over $7$ datasets.}\label{tab:dgi_perf}
    \begin{tabular}{clrrrrrrrb}
    \toprule
                                        &Methods            & {\scriptsize\flickr}              & {\scriptsize\yelp}              & {\scriptsize\reddit}            & {\scriptsize\reddittwo}           & {\scriptsize\aproducts}           & {\scriptsize\ogbnarxiv}        & {\scriptsize\ogbnproducts}           &Avg. Rank \\\midrule
\multirow{3}{*}{\rotatebox{90}{SAGE}}   &Random             &\ms{53.72}{0.16}			        &\ms{63.03}{0.20}			      &\ms{96.50}{0.03}			        &\ms{51.76}{2.53}			        &\ms{77.58}{0.05}			        &\ms{72.00}{0.16}			     &\ms{80.05}{0.35}			            &3.00\\ 
                                        &DGI                &\bms{53.97}{0.13}			        &\ms{62.53}{0.31}			      &\ms{96.57}{0.03}			        &\ms{54.82}{1.42}			        &\ms{77.11}{0.08}			        &\ms{71.86}{0.33}			     &\bms{80.24}{0.57}			            &1.71\\ 
                                        &\mlpinit           &\ms{53.82}{0.13}			        &\bms{63.93}{0.23}			      &\bms{96.66}{0.04}			    &\bms{89.60}{1.60}			        &\bms{77.74}{0.06}			        &\bms{72.25}{0.30}			     &\ms{80.04}{0.62}			            &1.28\\ \midrule
                                        
\multirow{3}{*}{\rotatebox{90}{GCN}}    &Random             &\ms{50.90}{0.12}			        &\ms{40.08}{0.15}			      &\ms{92.78}{0.11}			        &\ms{27.87}{3.45}			        &\ms{36.35}{0.15}			        &\ms{70.25}{0.22}			     &\ms{77.08}{0.26}			            &3.00\\
                                        &DGI                &\bms{51.23}{0.07}			        &\ms{38.24}{0.54}			      &\bms{94.14}{0.02}			    &\ms{66.98}{1.22}			        &\ms{35.54}{0.05}			        &\ms{69.40}{0.35}			     &\bms{77.15}{0.21}			            &1.57\\ 
                                        &\mlpinit           &\ms{51.16}{0.20}			        &\bms{40.83}{0.27}			      &\ms{91.40}{0.20}			        &\bms{80.37}{2.61}			        &\bms{39.70}{0.11}			        &\bms{70.35}{0.34}			     &\ms{76.85}{0.34}			            &1.42\\
    \bottomrule
    \end{tabular}
\end{table}

\begin{table}[t]
\setlength{\tabcolsep}{16pt}
\centering
\caption{The comparison of the running time of \mlpinit and DGI. We report the running time of one epoch for MLPInit and DGI in the pretraining stage and their ratio. The results show that training MLP is much cheaper than DGI, especially for the large graph.}\label{tab:dgi_run_time}
\begin{tabular}{lrrb}
\toprule
Dataset             &  MLPInit(ours)          & DGI                   & MLPInit/DGI Ratio  \\ \midrule
\ogbnarxiv          & \ms{0.035}{0.000}       & \ms{4.794}{0.055}     & $\mathbf{1/137}$    \\
\ogbnproducts       & \ms{0.076}{0.000}       & \ms{1892.386}{46.176} & $\mathbf{1/24798}$  \\
\bottomrule
\end{tabular}
\end{table}

\subsection{Experiments on more complicated aggregators}\label{sec:app:addition:aggr}

Information aggregators play a vital role in graph neural networks, and recent work proposed complicated aggregators to improve the performance of graph neural networks. In this appendix, we conducted experiments to investigate the effectiveness of \mlpinit on GNN with more complicated aggregators. We explore the acceleration effect and prediction accuracy improvement of \mlpinit on GNN with more complicated aggregators. The adopted aggregators include Mean, Sum, Max, Median, and Softmax. Their details are presented as follows:

\begin{itemize}
    \item \textbf{Mean} is a commonly used aggregation operator that averages features across neighbors. 
    \item \textbf{Max, Median} \citep{corso2020principal} are aggregation operators that take the feature-wise maximum/median across neighbors. The mathematical expressions of Max, Median are $\max_{\mathbf{x}_i \in \mathcal{X}} \mathbf{x}_i.$  $\text{median}_{\mathbf{x}_i \in \mathcal{X}} \mathbf{x}_i.$ 
    \item \textbf{Softmax} \citep{li2020deepergcn} is a learnable aggregation operator, which normalizes the features of neighbors based on a learnable temperature term, as $\mathrm{softmax}(\mathcal{X}|t) = \sum_{\mathbf{x}_i\in\mathcal{X}}
\frac{\exp(t\cdot\mathbf{x}_i)}{\sum_{\mathbf{x}_j\in\mathcal{X}}
\exp(t\cdot\mathbf{x}_j)}\cdot\mathbf{x}_{i}$ where $t$ controls the softness of the softmax when aggregating neighbors' features.
\end{itemize}

\begin{figure}[t]
    \centering
    \includegraphics[width=0.95\textwidth]{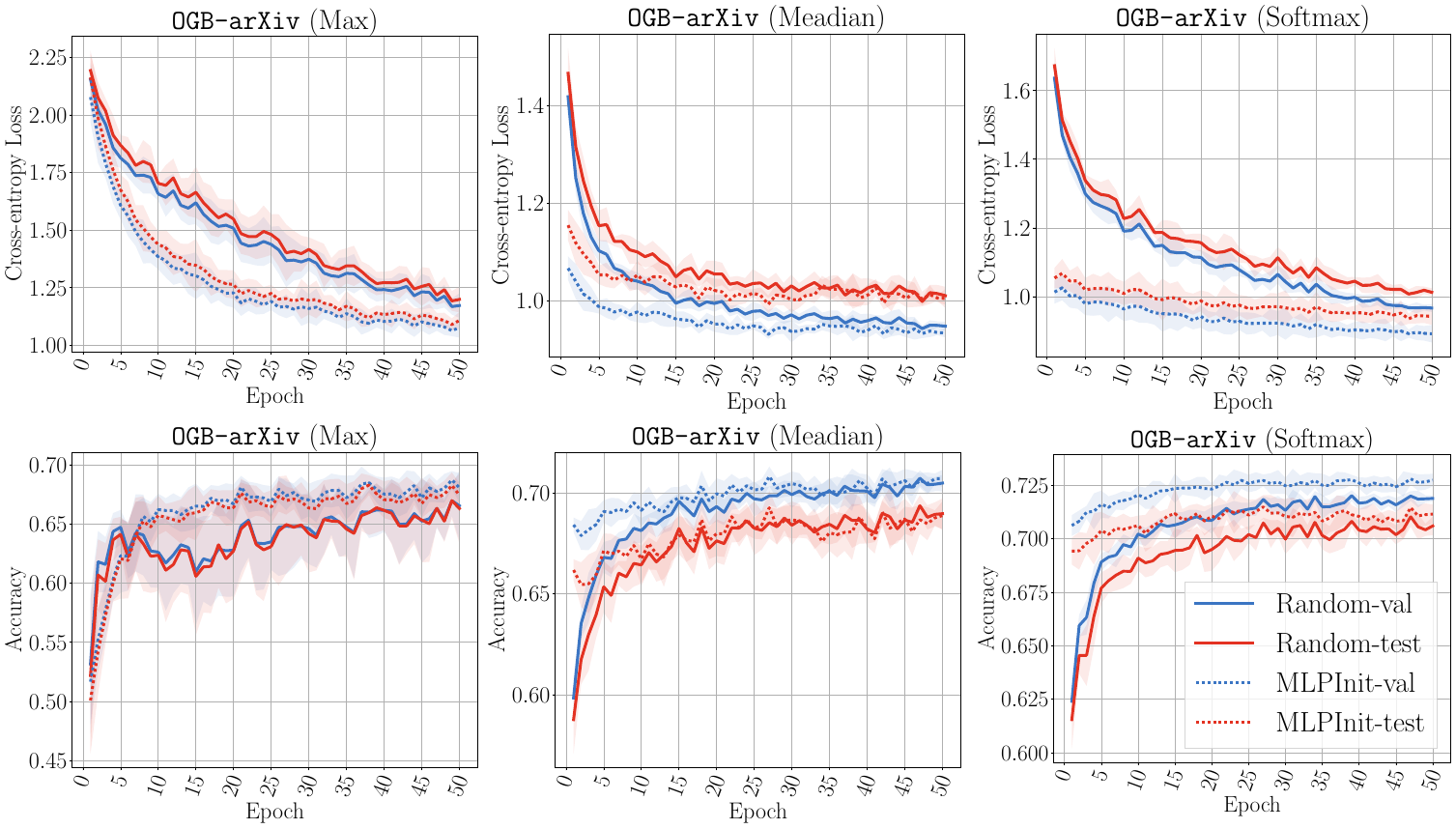}
    \caption{The training curves for GraphSAGE using different aggregators with random initialization and \mlpinit.}\label{fig:aggr_arxiv}
\end{figure}

We reported the performance improvement, training speedup, and training curves in \cref{tab:aggr_perf}, \cref{tab:aggr_speed} and \cref{fig:aggr_arxiv}. Generally, \mlpinit is effective for other aggregators. \mlpinit speed up the training of GNNs by $2.06\times$ over four aggregators. Moreover, \mlpinit improves the prediction performance by $0.75\%$ over four aggregators. The training curves in \cref{fig:aggr_arxiv} show that GNN with \mlpinit generally obtain lower loss and higher accuracy than those with random initialization and converge faster.

\begin{table}[t]
    \setlength{\tabcolsep}{11pt}
    \centering
    \caption{Performance improvement when GNN (GraphSAGE) uses different aggregators with random initialization. Mean, and standard deviation are calculated based on ten runs.}\label{tab:aggr_perf}
    \begin{tabular}{clrrrrb}
    \toprule
                                                   &Methods            & {Mean}                              &  {Max}                            & {Median}                           & {Softmax}                              &Avg. \\\midrule
\multirow{3}{*}{\rotatebox{90}{\tiny\ogbnarxiv}}   &Random             &\ms{72.00}{0.16}			         &\ms{68.31}{1.00}			         &\ms{69.97}{0.29}			          &\ms{71.05}{0.20}                        &70.33\\ 
                                                   &\mlpinit           &\ms{72.25}{0.30}			          &\ms{69.30}{0.56}			          &\ms{69.95}{0.36}		               &\ms{71.94}{0.18}	                    &70.86\\
                                                   &Improv.        &{\color{blue}$\uparrow 0.36\%$}     &{\color{blue}$\uparrow 1.44\%$}   &{\color{myred}$\downarrow 0.02\%$}    &{\color{blue}$\uparrow 1.25\%$}	    &{\color{blue}$\uparrow 0.75\%$}\\

    \bottomrule
    \end{tabular}
\end{table}

\begin{table}[H]
    \setlength{\tabcolsep}{15pt}
    \centering
    \caption{Speed improvement when {\mlpinit} achieves comparable performance with a randomly initialized GNN using various information aggregators. The number reported is the number of training epochs needed.}\label{tab:aggr_speed}
    \begin{tabular}{clrrrrb}
    \toprule
                                                   &Methods            & {Mean}                              & {Max}                              & {Median}                              & {Softmax}                               &Avg. \\\midrule
\multirow{3}{*}{\rotatebox{90}{\tiny\ogbnarxiv}}   &Random             &46.7			                     &37.1			                      &40.9			                          &42.0	                                    &41.6\\ 
                                                   &\mlpinit           &22.7			                      &22.4			                       &27.2			                       &8.8	                                     &20.2\\
                                                   &Improv.           &{\color{blue}2.06$\times$}          &{\color{blue}1.66$\times$}         &{\color{blue}1.50$\times$}            &{\color{blue}4.77$\times$}	             &{\color{blue}2.06$\times$} \\

    \bottomrule
    \end{tabular}
\end{table}

\subsection{Results on graph sampling methods and GNN architectures}
In this appendix, we provided the new form of our results in \cref{tab:perf} to show the performance improvements for graph sampling methods and GNN architectures separately in \cref{tab:perf_gsampling,tab:perf_gnnarch}.

From the new form of our results, we observed that 1) \mlpinit improves the performance of different graph sampling methods as it improves the performance of GraphSAGE, GraphSAINT, and ClusterGCN by $7.96\%$, $7.00\%$ and $6.62\%$. 2) \mlpinit improves the performance of different graph neural network layers, as it improves the performance of SAGEConv, GCNConv by $6.62\%$, $14.00\%$.

\begin{table}[t]
\centering
\setlength{\tabcolsep}{2pt}
\caption{Performance improvements for various graph sampling methods.  The improvement is the overall accuracy performance improvement across all the datasets.}\label{tab:perf_gsampling}

\scalebox{0.88}{
    \begin{tabular}{lccccccccc}
    \toprule
    Sampling & Methods & \flickr & \yelp & \reddit & \reddittwo & \aproducts & \ogbnarxiv & \ogbnproducts & improv. \\ \midrule
    \multirow{2}{*}{GraphSAGE}  & Random             & $53.72$             & $63.03$           & $96.50$             & $51.76$              & $77.58$                 & $72.00$                & $80.05$              &{\color{blue}\multirow{2}{*}{$\uparrow 7.96\%$}}               \\
                        & \mlpinit            & $53.82$             & $63.93$           & $96.66$             & $89.60$              & $77.74$                 & $72.25$                & $80.04$                   &        \\ \midrule
    \multirow{2}{*}{GraphSAINT}          & Random             & $51.37$             & $29.42$           & $95.58$             & $36.45$              & $59.31$                 & $67.95$                & $73.80$                   &{\color{blue}\multirow{2}{*}{$\uparrow 7.00\%$ }}              \\
                        & \mlpinit            & $51.35$             & $43.10$           & $95.64$             & $41.71$              & $68.24$                 & $68.80$                & $74.02$                   &          \\\midrule
    \multirow{2}{*}{Cluster-GCN}         & Random             & $49.95$             & $56.39$           & $95.70$             & $53.79$              & $52.74$                 & $68.00$                & $78.71$                   &{\color{blue}\multirow{2}{*}{$\uparrow 6.62\%$ }}             \\
                        & \mlpinit            & $49.96$             & $58.05$           & $96.02$             & $77.77$              & $55.61$                 & $69.53$                & $78.48$                   &           \\ 
    \bottomrule
    \end{tabular}
}

\end{table}

\begin{table}[t]
\centering
\setlength{\tabcolsep}{2pt}
\caption{Performance improvements for various GNN architectures. The improvement is the overall accuracy performance improvement across all the datasets.}\label{tab:perf_gnnarch}
\scalebox{0.88}{
    \begin{tabular}{lccccccccc}
    \toprule
    GNN layers& Methods & $\flickr$ & $\yelp$ & $\reddit$ & $\reddittwo$ & $\aproducts$ & $\ogbnarxiv$ & $\ogbnproducts$ & \textbf{ Improv. }  \\ \midrule
    \multirow{2}{*}{SAGEConv}           & Random             & $49.95$             & $56.39$           & $95.70$             & $53.79$              & $52.74$                 & $68.00$                & $78.71$                   &$\color{blue}\multirow{2}{*}{$\uparrow 6.62\%$} $             \\
                                        & \mlpinit            & $49.96$             & $58.05$           & $96.02$             & $77.77$              & $55.61$                 & $69.53$                & $78.48$                   &          \\\midrule
    \multirow{2}{*}{GCNConv}            & Random             & $50.90$             & $40.08$           & $92.78$             & $27.87$              & $36.35$                 & $70.25$                & $77.08$                   &$\color{blue}\multirow{2}{*}{$\uparrow 14.00\%$} $             \\
                                        & \mlpinit            & $51.16$             & $40.83$           & $91.40$             & $80.37$              & $39.70$                 & $70.35$                & $76.85$                   &          \\
    \bottomrule
    \end{tabular}
}

\end{table}

\subsection{Experiments on datasets where node features are less important}
Our proposed method does depend on a tendency towards node feature-label correlation. Thus, it would likely suffer if features provided less or no information about labels. In this appendix, we conducted experiments on synthetic graphs. The synthetic graphs have differing degrees of correlation between node features and labels. We generated the synthetic graph node features $\mathbf{X}_{synthetic}$ by mixing the original features and random features for each node in the graph as follows:

$$\mathbf{X}_{synthetic}=\lambda\mathbf{X}_{original}+(1-\lambda)\mathbf{X}_{random}$$

where $\mathbf{X}_{original}$ and $\mathbf{X}_{random}$ are the original features of \ogbnarxiv and random features, and $\lambda$ mediates the two. $\lambda$ shows different levels of association between node features and labels. When $\lambda=0$, the synthetic node features will be the original features of \ogbnarxiv. When $\lambda=1$, the synthetic node features will be completely random features, which are totally uncorrelated to the node labels. We change the value of $\lambda$ to explore the behavior of \mlpinit.  Note that we initially conducted the experiments on $\lambda=[0.0,0.1,0.2,...,1.0]$, and we observed that performance on $\lambda=0.1$ is much lower than other values, thus we conducted more on $\lambda=[0.05, 0.15]$ around $0.1$.

\begin{table}[t]
    \centering
    \setlength{\tabcolsep}{1.2pt}
    \caption{Performance of \peermlp and GNN with converged weights ($w^*_{mlp}$) of \peermlp on different $\lambda$ on dataset \ogbnarxiv. The accuracy in percentage is based on $5$ runs.}\label{tab:my_label}
\scalebox{0.77}{
    \begin{tabular}{l|cccccccccccccccc}
    \toprule
    \textbf{ $\lambda$ } & \textbf{ 0.0 } & \textbf{(0.05) } & \textbf{ 0.1 } & \textbf{ (0.15)} & \textbf{ 0.2 } & \textbf{ 0.3 } & \textbf{ 0.4 } & \textbf{ 0.5 } & \textbf{ 0.6 } & \textbf{ 0.7 } & \textbf{ 0.8 } & \textbf{ 0.9 } & \textbf{ 1.0 } \\
    \midrule
    \peermlp & $5.95$ & $6.63$ & $11.88$ & $16.39$ & $19.82$ & $23.59$ & $25.63$ & $31.19$ & $40.65$ & $48.74$ & $53.40$ & $55.67$ & $56.23$ \\
    GNN w/ $w^*_{mlp}$ & $5.86$ & $5.86$ & $5.89$ & $8.82$ & $24.10$ & $32.87$ & $40.67$ & $51.41$ & $55.99$ & $59.05$ & $61.30$ & $62.14$ & $62.43$ \\
    \midrule
    Improv. & $-1.53\%$ & $-11.59\%$ & $-50.43\%$ & $-46.16\%$ & $21.58\%$ & $39.37\%$ & $58.71\%$ & $64.81\%$ & $37.73\%$ & $21.15\%$ & $14.79\%$ & $11.63\%$ & $11.03\%$ \\
    \bottomrule
    \end{tabular}
}

\end{table}

\begin{table}
\centering
\setlength{\tabcolsep}{1.2pt}
\caption{Performance of GNN trained with Random Init and \mlpinit with different $\lambda$ on \ogbnarxiv. The accuracy in percentage is the best performance of the two methods and is based on $5$ runs.}\label{tab:my_label2}
\scalebox{0.77}{
    \begin{tabular}{l|ccccccccccccc|b}
    \toprule
    \textbf{ $\lambda$ } & \textbf{ 0.0 } & \textbf{(0.05) } & \textbf{ 0.1 } & \textbf{ (0.15)} & \textbf{ 0.2 } & \textbf{ 0.3 } & \textbf{ 0.4 } & \textbf{ 0.5 } & \textbf{ 0.6 } & \textbf{ 0.7 } & \textbf{ 0.8 } & \textbf{ 0.9 } & \textbf{ 1.0 } & \textbf{ Avg. }  \\ \midrule
    GNN w/ Random Init  & $60.03$          & $59.22$            & $60.71$          & $63.24$            & $64.71$          & $67.46$          & $69.02$          & $70.05$          & $70.72$          & $71.31$          & $71.76$          & $71.70$          & $72.00$          & $67.07$            \\
    GNN w/ \mlpinit  & $59.78$          & $60.47$            & $62.53$          & $64.49$            & $65.94$          & $67.84$          & $69.11$          & $70.47$          & $70.70$          & $71.45$          & $71.50$          & $71.92$          & $72.25$          & $67.57$            \\ \midrule
    Improv.                & $-0.42\%$        & $2.12\%$           & $3.01\%$         & $1.98\%$           & $1.91\%$         & $0.57\%$         & $0.13\%$         & $0.59\%$         & $-0.03\%$        & $0.20\%$         & $-0.36\%$        & $0.29\%$         & $0.34\%$         & $0.74\%$     \\     
    \bottomrule
    \end{tabular}
}

\end{table}

The results shows that
\begin{itemize}
    \item If node features are uncorrelated to the node labels ($\lambda < 0.2$), GNN with the weights of \peermlp will not outperform the \peermlp.
    \item If the node features are correlated to the node labels ($\lambda > 0.2$), GNN with the weights of \peermlp will consistently outperform the \peermlp.
    \item Overall, \mlpinit obtains a better ($0.74\%$ average improvement) final accuracy than Random Init over $13$ different $\lambda$s.
\end{itemize}

\subsection{Deriving the \peermlp}
In this appendix, we discuss two potential methods to derive the \peermlp, and discuss their advantages and disadvantages.

\subsubsection{Two Methods to Derive the \peermlp}
The two potential methods are as follows:

\begin{enumerate}
    \item \textbf{Remove the information aggregation operation in GNN}. In this way, we construct a new neural network (\peermlp, which may contain skip-connections or other complexities of the GNN layer) by entirely removing the neighbor aggregation operation; hence, the trainable weights of \peermlp will be the same as GNN by design. We need to build a dataloader for it (Algorithm 1). The advantage of this strategy is that it is efficient, since the \peermlp is a "pure" MLP (no aggregation required by design).
    \item \textbf{Change the adjacency matrix to an identity matrix} In this way, we use the original GNN architecture, but pretend the set of edges is a set of self-loops on each of the nodes.  The advantage of this strategy is that the same dataloader and model structure for GNN can be used for MLP -- we don't need to change the input of \peermlp, which are node features and adjacency matrix (changed to an identity matrix). This facilitates code reuse and ease of engineering/development.  However, since we also must use the GNN dataloader and associated model forward operations, we pay for some more training time owing to these operations (graph sampling and identity aggregation).
\end{enumerate}

\begin{table}[H]
\centering
\setlength{\tabcolsep}{2.0pt}
\caption{Comparison of the running time of forward and backward for different operations (i.e., $\mathbf{X} \cdot \mathbf{W}$, $\mathbf{A} \cdot \mathbf{Z}$ and $\mathbf{I} \cdot \mathbf{Z}$) in GNNs. The time unit is milliseconds (ms).}\label{fig:ix}

\scalebox{0.81}{
    \begin{tabular}{crrrrrrrrrrrrr}
    \toprule
    & \textbf{ } & \ogbnarxiv & \textbf{ } & \textbf{ } & \flickr& \textbf{ } & \textbf{ } & \yelp & \textbf{ }  \\ \midrule
    \#Nodes & & 169343 & & & 89250 & & & 716847 & \\
    \#Edges & & 1166243 & & & 899756 & & & 13954819 & \\\midrule
    & Forward & Backward & Total & Forward & Backward & Total & Forward & Backward & Total \\\midrule
    $\mathbf{Z} = \mathbf{X} \cdot \mathbf{W}$ & $0.56$ & $1.31$ & $1.87$ & $0.51$ & $1.80$ & $2.31$ & $3.10$ & $9.14$ & $12.24$ \\
    $\mathbf{H} = \mathbf{A} \cdot \mathbf{Z}$ & $1.47$ & $1285.86$ & $1287.33$ & $1.16$ & $842.49$ & $843.65$ & $13.19$ & $16956.19$ & $16969.38$ \\
    $\mathbf{H} = \mathbf{I} \cdot \mathbf{Z}$ & $1.32$ & $112.82$ & $114.13$ & $0.70$ & $63.55$ & $64.25$ & $5.61$ & $440.51$ & $446.12$ \\
    \bottomrule
    \end{tabular}
}

\end{table}

\begin{table}[H]
\centering
\setlength{\tabcolsep}{2.0pt}
\caption{Comparison of the running time of forward and backward for different methods (i.e., $\mathbf{W} \cdot \mathbf{X}$, $\mathbf{A} \cdot \mathbf{W} \cdot \mathbf{X}$ and $\mathbf{I} \cdot \mathbf{W} \cdot \mathbf{X}$). The time unit is milliseconds (ms).}\label{fig:ixx}

\scalebox{0.81}{
    \begin{tabular}{crrrrrrrrrrrrr}
    \toprule
    \textbf{}                                                                              & \textbf{ } & \ogbnarxiv & \textbf{ } & \textbf{ } & \flickr & \textbf{ } & \textbf{ } & \yelp & \textbf{ }  \\ \midrule
                                                                                                    & Forward    & Backward             & Total      & Forward    & Backward          & Total      & Forward    & Backward        & Total       \\
    $\mathbf{X} \cdot \mathbf{W}$                                                                & $0.56$       & $1.31$                 & $1.87$       & $0.51$       & $1.80$              & $2.31$       & $3.10$       & $9.14$            & $12.24$       \\
    $\mathbf{A} \cdot \mathbf{X} \cdot \mathbf{W}$                                               & $2.03$       & $1287.17$              & $1289.2$     & $1.67$       & $844.29$            & $845.96$     & $16.29$      & $16965.33$        & $16981.62$    \\
    $\mathbf{I} \cdot \mathbf{X} \cdot \mathbf{W}$                                               & $1.88$       & $114.13$               & $116$        & $1.21$       & $65.35$             & $66.56$      & $8.71$       & $449.65$          & $458.36$      \\ \midrule
    ratio of ($\mathbf{X} \cdot \mathbf{W}$):($\mathbf{A} \cdot \mathbf{X} \cdot \mathbf{W}$) & ---        & ---                  & $1{:}689$  & ---        & ---               & $1{:}366$  & ---        & ---             & $1{:}1387$  \\
    ratio of ($\mathbf{X} \cdot \mathbf{W}$):($\mathbf{I} \cdot \mathbf{X} \cdot \mathbf{W}$) & ---        & ---                  & $1{:}62$   & ---        & ---               & $1{:}28$   & ---        & ---             & $1{:}37$   \\
    \bottomrule
    \end{tabular}
}

\end{table}

For more complex graph convolution layers, we take a layer with skip-connection in GNN, $\mathbf{H}^{l} = \sigma( \mathbf{A} \mathbf{H}^{l-1} \mathbf{W}_{gnn}^{l} + \mathbf{H}^{l-1})$, as an example. To derive the \peermlp for the layer with skip-connection, Method 1 directly removes the adjacency matrix $\mathbf{A}$ to yield $\mathbf{H}^{l} = \sigma( \mathbf{H}^{l-1} \mathbf{W}_{gnn}^{l} + \mathbf{H}^{l-1})$. Thus, the \peermlp will also contain a skip-connection operation. Method 2 can be easily and directly adopted since it just alters the adjacency matrix in a trivial way. $\mathbf{H}^{l} = \sigma(\mathbf{H}^{l-1} \cdot \mathbf{W}_{gnn}^{l} + \mathbf{H}^{l-1})$.

\subsubsection{Discussion about the efficiency of \peermlp deriving}

Firstly, Method 1 and Method 2 are mathematically equivalent.

Secondly, in the sense of producing computational graphs, they are different. In the next, we only consider the terms $\mathbf{H}^{l-1} \cdot \mathbf{W}_{gnn}^{l}$ and $\mathbf{I} \cdot \mathbf{H}^{l-1} \cdot \mathbf{W}_{gnn}^{l}$ in Method 1 and 2 since the rest terms are the same. For operation $\mathbf{W}_{gnn}^{l} \cdot \mathbf{H}^{l-1}$ in Method 1, it only has one dense matrix multiplication. For operation $\mathbf{I} \cdot \mathbf{H}^{l-1} \cdot \mathbf{W}_{gnn}^{l}$ in Method 2, it has two steps, one is dense matrix multiplication ($\mathbf{Z} = \mathbf{H}^{l-1} \cdot \mathbf{W}_{gnn}^{l}$), the other is a sparse matrix multiplication ($\mathbf{I} \cdot \mathbf{Z}$). Thus they will produce different computational graphs (using torch\_geometric). Typically, sparse matrix multiplication needs much more time than dense matrix multiplication (in the case that the sparse matrix is an identity matrix, this still holds if the computation package has no special optimization for the identity matrix).

To investigate the efficiency of these two methods in production environments, we conducted experiments to present the running time of different operations ($\mathbf{X}  \cdot \mathbf{W}$, $\mathbf{A} \cdot \mathbf{Z}$ and $\mathbf{I} \cdot \mathbf{Z}$) and different methods ($\mathbf{X} \cdot \mathbf{W}$, $\mathbf{A} \cdot \mathbf{X} \cdot \mathbf{W}$ and $\mathbf{I} \cdot \mathbf{X} \cdot \mathbf{W}$). $\mathbf{X} \cdot \mathbf{W}$ is dense matrix multiplication, $\mathbf{A} \cdot \mathbf{Z}$ and $\mathbf{I} \cdot \mathbf{Z}$ are sparse matrix multiplication. The experiments are conducted with an NVIDIA RTX A5000. The softwares and their version in this experiment are cudatoolkit (11.0.221), PyTorch (1.9.1), torch-sparse (0.6.12). The results are presented in \cref{fig:ix,fig:ixx}. From the results, we have the following observation:

\begin{itemize}
    \item \cref{fig:ix} shows that $\mathbf{I} \cdot \mathbf{Z}$ (sparse matrix multiplication)  takes much more time than $\mathbf{X} \cdot \mathbf{W}$ (dense matrix multiplication).
    \item \cref{fig:ixx} shows that $\mathbf{I} \cdot \mathbf{X} \cdot \mathbf{W}$ takes much more time than $\mathbf{X} \cdot \mathbf{W}$, where the ratio of the running time for ($\mathbf{X} \cdot \mathbf{W}$):($\mathbf{I} \cdot \mathbf{X} \cdot \mathbf{W}$) are $1:62$, $1:28$, and $1:37$ for \ogbnarxiv, \flickr and \yelp, respectively. The results indicate that the $\mathbf{I} \cdot \mathbf{X} \cdot \mathbf{W}$ is not equivalent to $\mathbf{X} \cdot \mathbf{W}$ in the sense of producing computational graph (at least for the sparse matrix multiplication in torch-sparse package).
    \item The choice between the two ultimately boils down to the intended setting.  Method 2 is easier for development, and Method 1 is more optimized for speed (and thus useful in resource-constrained settings, or for efficiency in production environments).
\end{itemize}

\clearpage
\section{Datasets and Baselines}\label{sec:app:setting}
In the appendix, we present the details of datasets and baselines for node classification and link prediction tasks.

\subsection{Datasets for node classification}\label{sec:app:setting:data}
The details of datasets used for node classification are listed as follows:
\begin{itemize}
\item \yelp \citep{zeng2019graphsaint} contains customer reviewers as nodes and their friendship as edges. The node features are the low-dimensional review representations for their reviews.
\item \flickr \citep{zeng2019graphsaint} contains customer reviewers as nodes and their common properties as edges. The node features the 500-dimensional bag-of-word representation of the images.
\item \reddit, \reddittwo \citep{hamilton2017inductive} is constructed by Reddit posts. The node in this dataset is a post belonging to different communities. \reddittwo is the sparser version of \reddit by deleting some edges.
\item \aproducts \citep{zeng2019graphsaint} contains products and its categories.
\item \ogbnarxiv \citep{hu2020open} is the citation network between all arXiv papers. Each node denotes a paper and each edge denotes citation between two papers. The node features are the average 128-dimensional word vector of its title and abstract.
\item \ogbnproducts \citep{hu2020open,chiang2019cluster} is Amazon product co-purchasing network. Nodes represent products in Amazon, and edges between two products indicate that the products are purchased together. Node features in this dataset are low-dimensional representations of the product description text.
\end{itemize}
We present the statistics of datasets used for node classification task in \cref{tab:dataset:nc}.

\begin{table}[!htb]
    \centering
    \caption{Statistics for datasets used for node classification.}
    \label{tab:dataset:nc}
    {\footnotesize
    \begin{tabular}{crrrrrr}
    \toprule
         Dataset                    & \# Nodes. & \# Edges      &\# Classes     &\# Feat          &Density \\\midrule
         \flickr            &89,250     &899,756        &7              &500              &0.11\textperthousand  \\
         \yelp              &716,847    &13,954,819     &100            &300              &0.03\textperthousand  \\
         \reddit            &232,965    &114,615,892    &41             &602              &2.11\textperthousand  \\
         \reddittwo         &232,965    &23,213,838     &41             &602              &0.43\textperthousand  \\
         \aproducts         &1,569,960  &264,339,468    &107            &200              &0.11\textperthousand  \\
         \ogbnarxiv         &169,343    &1,166,243      &40             &128              &0.04\textperthousand  \\
         \ogbnproducts      &2,449,029  &61,859,140     &47             &218              &0.01\textperthousand  \\
    \bottomrule
    \end{tabular}
    }
\end{table}

\subsection{Baselines for node classification}\label{sec:app:setting:base}
We present the details of GNN models as follows:
\begin{itemize}
    \item \textbf{GCN}~\citep{kipf2016semi,hamilton2017inductive} is the original graph convolution network, which aggregates neighbor's information to obtain node representation. In our experiment, we train GCN in a mini-batch fashion by adopting a subgraph sampler from \cite{hamilton2017inductive}.
    \item \textbf{GraphSAGE}~\citep{hamilton2017inductive} proposes a graph sampling-based training strategy to scale up graph neural networks. It samples a fixed number of neighbors per node and trains the GNNs in a mini-batch fashion.
    \item \textbf{GraphSAINT}~\citep{zeng2019graphsaint} is a graph sampling-based method to train GNNs on large-scale graphs, which proposes a set of graph sampling algorithms to partition graph data into subgraphs. This method also presents normalization techniques to eliminate biases during graph sampling.
    \item \textbf{ClusterGCN}~\citep{chiang2019cluster} is proposed to train GCNs in small batches by using the graph clustering structure. This approach samples the nodes associated with the dense subgraphs identified by the graph clustering algorithm. Then the GCN is trained by the subgraphs.
\end{itemize}

\subsection{Datasets for link prediciton}\label{sec:app:setting:data:lp} 

For link prediction task, we consider \cora, \seer, \pubmed, \corafull, \cs, \physics, \photo. and \computers as our baselines. The details of the datasets used for node classification are listed as follows:

\begin{itemize}
\item \cora, \seer, \pubmed~\citep{yang2016revisiting} are representative citation network datasets. These datasets contain a number of research papers, where nodes and edges denote documents and citation relationships, respectively. Node features are low-dimension representations for papers. Labels indicate the research field of documents.

\item \corafull \citep{bojchevski2018deep} is a citation network that contains papers and their citation relationship. Labels are generated based on topics. This dataset is the original data of the entire network of Cora, and Cora dataset in Planetoid is its subset.

\item \cs, \physics~\citep{shchur2018pitfalls} are from Co-author dataset, which is co-authorship graph based on the Microsoft Academic Graph from the KDD Cup 2016 challenge. Nodes in this dataset are authors, and edges indicate co-author relationships. Node features represent paper keywords. Labels indicate the research field of the authors. 

\item \photo, \computers~\citep{shchur2018pitfalls} are two datasets from Amazon co-purchase dataset~\citep{mcauley2015image}. Nodes in this dataset represent products, while edges represent the co-purchase relationship between two products. Node features are low-dimension representations of product reviews. Labels are categories of products.
\end{itemize}
We also present the statistics of datasets used for link prediction task in \cref{tab:dataset:lp}.

\begin{table}[!htb]
    \centering
    \caption{Statistics for datasets used for link prediction.}\label{tab:dataset:lp}
    \begin{tabular}{crrrrrr}
    \toprule
         Dataset            & \# Nodes  & \# Edges      &\# Feat    &Density  \\\midrule
         \cora              &2,708      &5,278          &1,433      &0.72\textperthousand   \\
         \seer              &3,327      &4,552          &3,703      &0.41\textperthousand   \\
         \pubmed            &19,717     &44,324         &500        &0.11\textperthousand   \\
         \dblp              &17,716     &105,734        &1,639      &0.34\textperthousand   \\
         \corafull          &19,793     &126,842        &8,710      &0.32\textperthousand   \\
         \photo             &7,650      &238,162        &745        &4.07\textperthousand   \\
         \computers         &13,752     &491,722        &767        &2.60\textperthousand   \\
         \cs                &18,333     &163,788        &6,805      &0.49\textperthousand   \\
         \physics           &34,493     &495,924        &8,415      &0.14\textperthousand   \\
    \bottomrule
    \end{tabular}
    
\end{table}

\subsection{Baselines for link prediciton}\label{sec:app:setting:base:lp}
Our link prediction setup is consistent with our discussion in \cref{sec:pre}, in which we use an inner-product decoder $\hat{\mathbf{A}} = \mathrm{sigmoid}( \mathbf{H}\cdot \mathbf{H}^T )$ to predict the probability of the link existence. We presented the results in \cref{tab:lp,tab:lp_app}. Following standard benchmarks~\citep{hu2020open}, the evaluation metrics adopted are AUC, Average Precision (AP), are hits ratio (Hit@\#). The experimental details of link prediction are presented in \cref{sec:app:imple:tab:lp}.

\clearpage
\section{Implementation Details}\label{sec:app:imple}

In this appendix, we present the hyperparameters used for the node classification task and link prediction task for all models and datasets.

\subsection{Running environment}\label{sec:app:setting:runing}
We run our experiments on the machine with one NVIDIA Tesla T4 GPU (16GB memory) and 60GB DDR4 memory to train the models. For \aproducts and \ogbnproducts datasets, we run the experiments with one NVIDIA A100 GPU (40GB memory). The code is implemented based on PyTorch 1.9.0~\citep{paszke2019pytorch} and PyTorch Geometric 2.0.4~\citep{fey2019fast}. The optimizer is Adam~\citep{kingma2015adam} to train all GNNs and their {\peermlp}s. 


\subsection{Experiment setting for \texorpdfstring{\cref{fig:gnn_mlp_loss}}{} }\label{sec:app:imple:fig:gnn_mlp_loss}
In this experiment, we use GraphSAGE as GNN on \ogbnarxiv dataset. We construct that {\peermlp}($f_{mlp}(\mathbf{X}; w_{mlp})$ ) for GraphSAGE ($f_{gnn}(\mathbf{X}, \mathbf{A}; w_{mlp})$) and train it for $50$ epochs. From the mathematical expression, GraphSAGE and its \peermlp share the same weights $w_{mlp}$ and the weights $w_{mlp}$ are only trained by \peermlp. We use the trained weights for GraphSAGE to make inference along the training procedure. For the landscape, suppose we have two optimal weight $w_{gnn}^{*}$ and $w_{mlp}^{*}$ for GraphSAGE and its  \peermlp, the middle one is the loss landscape based on the  \peermlp with optimal with $w_{mlp}^{*}$ while the right one is the loss landscape based the GraphSAGE with optimal weights $w_{gnn}^{*}$.

\subsection{Experiment setting for \texorpdfstring{\cref{fig:ogbn_arxiv_perf,fig:perf_add1,fig:perf_add2,tab:speed,tab:perf}}{} }\label{sec:app:imple:fig:ogbn_products_perf}
In this appendix, we present the detailed experiment setting for our main result \cref{fig:ogbn_arxiv_perf,fig:perf_add1,fig:perf_add2,tab:speed,tab:perf}. We construct the \peermlp for each GNN. We first train the  \peermlp for $50$ epochs and save the best model with the best validation performance. And then, we use the weight trained by  \peermlp to initialize the GNNs, then fine-tune the GNNs. To investigate the performance of GNNs, we fine-tune the GNNs for $50$ epochs. We list the hyperparameters used in our experiments. We borrow the optimal hyperparameters from paper \citep{duancomprehensive}. And our code is also based on the official code~\footnote{\url{https://github.com/VITA-Group/Large_Scale_GCN_Benchmarking}} of paper \citep{duancomprehensive}. For datasets not included in paper \citep{duancomprehensive}, we use the heuristic hyperparameter setting for them.

\begin{table}[t]
    \caption{Training configuration for GNNs training in \cref{fig:ogbn_arxiv_perf,fig:perf_add1,fig:perf_add2,tab:speed,tab:perf}. }\label{tab: config_nc}
        \centering
        \setlength{\tabcolsep}{2pt}
        \begin{tabular}{l|crrrrrrrr}
            \toprule
                Model               & Dataset           &\#Layers   & \#Hidden      &Learning rate  &Batch size  &Dropout   &Weight decay   &Epoch\\\midrule
    \multirow{7}{*}{\rotatebox{90}{GraphSAGE}}      &\flickr         	&$4$ 	        &$512$ 	        &$0.0001$ 	        &$1000$ 	     &$0.5$       &$0.0001$ 	    &$50$ \\
                                                    &\yelp              &$4$ 	        &$512$ 	        &$0.0001$ 	        &$1000$ 	     &$0.2$       &$0 	   $         &$50$ \\
                                                    &\reddit            &$4$ 	        &$512$ 	        &$0.0001$ 	        &$1000$ 	     &$0.2$       &$0 	   $         &$50$ \\
                                                    &\reddittwo         &$4$ 	        &$512$ 	        &$0.0001$ 	        &$1000$ 	     &$0.2$       &$0 	   $         &$50$ \\
                                                    &\aproducts         &$4$ 	        &$512$ 	        &$0.001 $	        &$1000$ 	     &$0.5$       &$0 	   $         &$50$ \\
                                                    &\ogbnarxiv         &$4$ 	        &$512$ 	        &$0.001 $	        &$1000$ 	     &$0.5$       &$0 	   $         &$50$ \\
                                                    &\ogbnproducts      &$4$ 	        &$512$ 	        &$0.001 $	        &$1000$ 	     &$0.5$       &$0 	   $         &$50$ \\\midrule
    \multirow{7}{*}{\rotatebox{90}{GraphSAINT}}     &\flickr         	&$4$ 	        &$512$ 	        &$0.001 $	        &$5000$ 	     &$0.2$       &$0.0004$ 	    &$50$ \\
                                                    &\yelp           	&$2$ 	        &$128$ 	        &$0.01 	$           &$5000$ 	     &$0.7$       &$0.0002$ 	    &$50$ \\
                                                    &\reddit         	&$2$ 	        &$128$ 	        &$0.01 	$           &$5000$ 	     &$0.7$       &$0.0002$ 	    &$50$ \\
                                                    &\reddittwo      	&$2$ 	        &$128$ 	        &$0.01 	$           &$5000$ 	     &$0.7$       &$0.0002$ 	    &$50$ \\
                                                    &\aproducts      	&$2$ 	        &$128$ 	        &$0.01 	$           &$5000$ 	     &$0.2$       &$0 	   $         &$50$ \\
                                                    &\ogbnarxiv      	&$2$ 	        &$128$ 	        &$0.01 	$           &$5000$ 	     &$0.2$       &$0 	   $         &$50$ \\
                                                    &\ogbnproducts   	&$2$ 	        &$128$ 	        &$0.01 	$           &$5000$ 	     &$0.2$       &$0 	   $         &$50$ \\\midrule
    \multirow{7}{*}{\rotatebox{90}{ClusterGCN}}     &\flickr         	&$2$ 	        &$256$ 	        &$0.001 $	        &$5000$ 	     &$0.2$       &$0.0002$ 	    &$50$ \\
                                                    &\yelp           	&$4$ 	        &$256$ 	        &$0.0001$ 	        &$2000$ 	     &$0.5$       &$0 	   $         &$50$ \\
                                                    &\reddit         	&$4$ 	        &$256$ 	        &$0.0001$ 	        &$2000$ 	     &$0.5$       &$0 	   $         &$50$ \\
                                                    &\reddittwo      	&$4$ 	        &$256$ 	        &$0.0001$ 	        &$2000$ 	     &$0.5$       &$0 	   $         &$50$ \\
                                                    &\aproducts      	&$4$ 	        &$128$ 	        &$0.001 $	        &$2000$ 	     &$0.2$       &$0.0001$ 	    &$50$ \\
                                                    &\ogbnarxiv      	&$4$ 	        &$128$ 	        &$0.001 $	        &$2000$ 	     &$0.2$       &$0.0001$ 	    &$50$ \\
                                                    &\ogbnproducts   	&$4$ 	        &$128$ 	        &$0.001 $	        &$2000$ 	     &$0.2$       &$0.0001$ 	    &$50$ \\\midrule
    \multirow{7}{*}{\rotatebox{90}{GCN}}            &\flickr         	&$2$ 	        &$512$ 	        &$0.0001$ 	        &$1000$ 	     &$0.5$       &$0.0001$ 	    &$50$ \\
                                                    &\yelp           	&$2$ 	        &$512$ 	        &$0.0001$ 	        &$1000$ 	     &$0.2$       &$0 	   $         &$50$ \\
                                                    &\reddit         	&$2$ 	        &$512$ 	        &$0.0001$ 	        &$1000$ 	     &$0.2$       &$0 	   $         &$50$ \\
                                                    &\reddittwo      	&$2$ 	        &$512$ 	        &$0.0001$ 	        &$1000$ 	     &$0.2$       &$0 	   $         &$50$ \\
                                                    &\aproducts      	&$2$ 	        &$512$ 	        &$0.001 $	        &$1000$ 	     &$0.5$       &$0 	   $         &$50$ \\
                                                    &\ogbnarxiv      	&$2$ 	        &$512$ 	        &$0.001 $	        &$1000$ 	     &$0.5$       &$0 	   $         &$50$ \\
                                                    &\ogbnproducts   	&$2$ 	        &$512$ 	        &$0.001 $	        &$1000$ 	     &$0.5$       &$0 	   $         &$50$ \\
            \bottomrule
        \end{tabular}
    
    \end{table}

\subsection{Experiment setting for \texorpdfstring{\cref{tab:gnn_mlp}}{} }\label{sec:app:imple:tab:gnn_mlp}
The GNN used in \cref{tab:gnn_mlp} is GraphSAGE. We construct the \peermlp for GraphSAGE on \ogbnarxiv and \ogbnproducts datasets. We first train the \peermlp for $50$ epochs and save the best model with the best validation performance. And then we infer the test performance with \peermlp and GraphSAGE with the weight trained by \peermlp and we report the test performance in \cref{tab:gnn_mlp}.

\subsection{Experiment setting for \texorpdfstring{\cref{tab:lp,tab:lp_app}}{} } \label{sec:app:imple:tab:lp}
In this appendix, we present the detailed experiment setting for the link prediction task. We adopt the default setting for the official examples~\footnote{ \url{https://github.com/pyg-team/pytorch_geometric/blob/2.0.4/examples/link_pred.py}} of PyTorch Geometric 2.0.4. The GNN used for the link prediction task is a 2-layer GCN, and the decoder is the commonly used inner-product decoder as $\hat{\mathbf{A}} = \mathrm{sigmoid}( \mathbf{H}\cdot \mathbf{H}^T )$~\citep{kipf2016variational}.

\subsection{Experiment setting for \texorpdfstring{\cref{fig:ogbn_arxiv_hps}}{} } \label{sec:app:imple:fig:ogbn_arxiv_hps}
We explore two kinds of hyperparameters  ``Training HP'' (Learning rate, weight decay, batch size, and dropout) and ``Architecture HP'' (i.e., layers, number of hidden neurons), in this experiment.

\begin{itemize}
    \item Training Hyperparameter ( Training HP ), total combinations: $2\times 2 \times 2 \times 2=16$
        \begin{itemize}
            \item Learning rate: $\{0.001, 0.0001\}$
            \item Weight decay: $\{1e-4, 4e-4\}$
            \item Batch size: $\{500, 1000\}$
            \item Dropout: $\{0.2, 0.5\}$
        \end{itemize}
    \item Architecture Hyperparameter ( Architecture HP ), total combinations: $3\times 5=15$
        \begin{itemize}
            \item Number of layers: $\{2, 3, 4 \}$
            \item number of hidden neurons: $\{32, 64, 128, 256, 512\}$
        \end{itemize}
\end{itemize}

In \cref{fig:ogbn_arxiv_hps,fig:ogbn_arxiv_hps_add}, we plotted the learning curves based on the mean and standard deviation over all the hyperparameters combinations.

\clearpage

\end{document}